\documentclass[10pt,twocolumn,letterpaper]{article}

\usepackage{cvpr}
\usepackage[utf8]{inputenc}
\usepackage{times}
\usepackage{epsfig}
\usepackage{graphicx}
\usepackage{amsmath,mathtools}
\usepackage{amssymb}
\usepackage{amsthm}
\usepackage{bm}
\usepackage{enumitem}
\usepackage{mathtools}
\usepackage{multirow}
\usepackage[htt]{hyphenat}
\usepackage{fontawesome}
\usepackage{pifont}
\usepackage{color}
\usepackage{xcolor}
\usepackage[algo2e,inoutnumbered,linesnumbered,algoruled,vlined]{algorithm2e}
\usepackage{algpseudocode}
\usepackage{booktabs}
\usepackage{tabularx}
\usepackage{bm}
\usepackage{url}
\usepackage{bbm}
\usepackage{threeparttable}
\usepackage{tikz}
\usepackage{subfig}
\usetikzlibrary{backgrounds}

\usepackage[pagebackref=true,breaklinks=true,letterpaper=true,colorlinks,bookmarks=false]{hyperref}

\newcommand{\parahead}[1]{\noindent\textbf{#1}:\ }

\DeclareMathOperator*{\argmin}{argmin}

\cvprfinalcopy %

\def\cvprPaperID{1285} %

\ifcvprfinal\fi

\newcommand{\x}{\mathbf{x}} %
\def\v{{\mathbf{v}}} %
\newcommand{\X}{\mathbf{X}} %
\newcommand{\Y}{\mathbf{Y}} %
\newcommand{\y}{\mathbf{y}} %
\newcommand{\V}{\mathbf{V}} %
\newcommand{\A}{\mathbf{A}} %
\newcommand{\C}{\mathbf{C}} %
\newcommand{\cx}{\mathbf{c}} %
\newcommand{\ab}{\mathbf{a}} %
\newcommand{\T}{\mathbf{T}} %
\newcommand{\tra}{\mathbf{t}} %
\newcommand{\Tego}{\T_{\mathrm{ego}}} %
\newcommand{\Tset}{\mathcal{T}} %
\newcommand{\M}{\mathbf{M}} %
\newcommand{\G}{\mathbf{G}} %
\newcommand{\F}{\mathbf{F}} %
\newcommand{\Edge}{\mathcal{E}} %
\newcommand{\h}{\mathbf{h}} %
\newcommand{\hgt}{\overbar{\h}} %
\newcommand{\Id}{\mathbf{I}} %
\newcommand{\R}{\mathbb{R}} %
\newcommand{\Rot}{\mathbf{R}} %
\newcommand{\Rego}{\mathbf{R}_\mathrm{ego}} %
\newcommand{\tego}{\mathbf{t}_\mathrm{ego}} %
\newcommand{\one}{\mathbf{1}} %
\newcommand{\zero}{\mathbf{0}} %
\newcommand{\SOg}{\mathrm{SO}(3)} %
\newcommand{\D}{\mathbf{D}} %

\newcommand{\loss}{\mathcal{L}} %
\newcommand{\clus}{\mathcal{C}} %
\newcommand{\flow}{\mathcal{V}} %
\newcommand{\pars}{\bm{\Gamma}}

\newcommand{\mur}{\bm{\mu}}
\newcommand{\muc}{\bm{\nu}}

\newcommand{\Pb}{\mathbf{P}}

\DeclarePairedDelimiterX{\infdivx}[2]{(}{)}{%
  #1\;\delimsize\|\;#2%
}

\newcommand{\overbar}[1]{\mkern 1.5mu\overline{\mkern-1.5mu#1\mkern-1.5mu}\mkern 1.5mu}

\newcommand{\inv}{^{\raisebox{.2ex}{$\scriptscriptstyle-1$}}}

\renewcommand{\parahead}[1]{\vspace{2mm}\noindent\textbf{#1}.\ }

\newcommand{\review}[1]{{\color{black} #1}}

\usepackage{scalerel,stackengine}
\def\apeqA{\SavedStyle\sim}
\def\apeq{\setstackgap{L}{\dimexpr.5pt+1.5\LMpt}\ensurestackMath{%
  \ThisStyle{\mathrel{\Centerstack{{\apeqA} {\apeqA} {\apeqA}}}}}}
  
\usepackage{cleveref}
\crefname{section}{\S}{\S\S}
\crefname{subsection}{\S}{\S\S}

\Crefname{assumption}{\textbf{H}\hspace{-3pt}}{\textbf{H}\hspace{-3pt}}
\crefname{assumption}{\textbf{H}}{\textbf{H}}

\crefname{algorithm}{\text{Alg.}}{\text{Alg.}}
\crefname{assumption}{\textbf{H}}{\textbf{H}}
\crefname{equation}{\text{Eq}}{\text{Eq}}
\crefname{definition}{\text{Dfn.}}{\text{Dfn.}}
\crefname{lemma}{\text{Lemma}}{\text{Lemma}}
\crefname{dfn}{\text{Dfn.}}{\text{Dfn.}}
\crefname{thm}{\text{Thm.}}{\text{Thm.}}
\crefname{tab}{\text{Tab.}}{\text{Tab.}}
\crefname{fig}{\text{Fig.}}{\text{Fig.}}
\crefname{table}{\text{Tab.}}{\text{Tab.}}
\crefname{figure}{\text{Fig.}}{\text{Fig.}}
\crefname{section}{\text{Sec.}}{\text{Sec.}}

\newcommand{\insertimageC}[5]{ %
\begin{figure}[#5]
\centering
\includegraphics[width=#1\linewidth, clip=true]{figures/#2}
\caption{#3}
\label{#4}
\end{figure}
}

\newcommand{\insertimageStar}[5]{ %
\begin{figure*}[#5]
\centering
\includegraphics[width=#1\linewidth, clip=true]{figures/#2}
\caption{#3}
\label{#4}
\end{figure*}
}

\begin{document}

\title{\vspace{-0.45cm}Weakly Supervised Learning of Rigid 3D Scene Flow\vspace{-0.3cm}}

\author{\vspace{0.2cm}Zan Gojcic$^{1,2}$ \qquad Or Litany $^{2,3}$ \qquad Andreas Wieser$^{1}$ \qquad Leonidas J. Guibas$^2$ \qquad Tolga Birdal$^2$ \\ \vspace{0.2cm}
$^{1}$ETH Zurich \qquad $^2$Stanford University \qquad $^3$NVIDIA \\ \url{3dsceneflow.github.io}}

\maketitle

\begin{abstract}

We propose a data-driven scene flow estimation algorithm exploiting the observation that many 3D scenes can be explained by a collection of agents moving as rigid bodies. At the core of our method lies a deep architecture able to reason at the \textbf{object-level} by considering 3D scene flow in conjunction with other 3D tasks. This object level abstraction, enables us to relax the requirement for dense scene flow supervision with simpler binary background segmentation mask and ego-motion annotations. Our mild supervision requirements make our method well suited for recently released massive data collections for autonomous driving, which do not contain dense scene flow annotations. As output, our model provides low-level cues like pointwise flow and higher-level cues such as holistic scene understanding at the level of rigid objects. We further propose a test-time optimization refining the predicted rigid scene flow. We showcase the effectiveness and generalization capacity of our method on four different autonomous driving datasets. We release our source code and pre-trained models under \url{github.com/zgojcic/Rigid3DSceneFlow}.

\vspace{-5mm}
\end{abstract}
\section{Introduction}\vspace{-1mm}
\label{sec:intro}

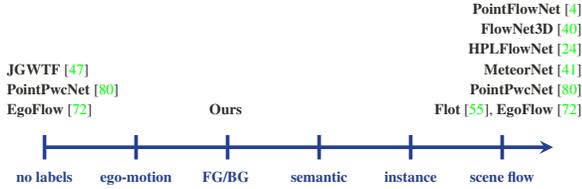
\begin{figure}
\centering
  \resizebox{\columnwidth}{!}{
  \tikzset{>=stealth}
\tikzset{every node/.append style={text depth=0.7ex}}
\definecolor{cardinal}{RGB}{37,64,143}
\definecolor{stanfordgrey}{RGB}{46,45,41}

\begin{tikzpicture}
\draw[line width=0.6 mm, ->, color=cardinal] (0,0) -- (\columnwidth,0);
\draw[line width=0.6 mm, color=cardinal] (0,-0.2) -- (0,0.2);
\draw[line width=0.6 mm, color=cardinal] (1.5,-0.2) -- (1.5,0.2);
\draw[line width=0.6 mm, color=cardinal] (3,-0.2) -- (3,0.2);
\draw[line width=0.6 mm, color=cardinal] (4.5,-0.2) -- (4.5,0.2);
\draw[line width=0.6 mm, color=cardinal] (6,-0.2) -- (6,0.2);
\draw[line width=0.6 mm, color=cardinal] (7.5,-0.2) -- (7.5,0.2);

\node[color=cardinal, align=center,anchor=south] at (0,-0.8) {\scriptsize \textbf{no labels}};
\node[color=cardinal, align=center,anchor=south] at (1.5,-0.8) {\scriptsize \textbf{ego-motion}};
\node[color=cardinal, align=center,anchor=south] at (3,-0.8) {\scriptsize \textbf{FG/BG} };
\node[color=cardinal, align=center,anchor=south] at (4.5,-0.8) {\scriptsize \textbf{semantic}};
\node[color=cardinal, align=center,anchor=south] at (6,-0.8) {\scriptsize \textbf{instance}};
\node[color=cardinal, align=center,anchor=south] at (7.5,-0.8) {\scriptsize \textbf{scene flow}};
{\scriptsize \textbf{instance}};

\node[color=stanfordgrey, anchor=south, align=right, execute at begin node=\setlength{\baselineskip}{2.1ex}] at (7.6,0.3) { \scriptsize \textbf{PointFlowNet}~\cite{behl2019pointflownet}\\
\scriptsize \textbf{FlowNet3D}~\cite{liu2019flownet3d}\\
\scriptsize \textbf{HPLFlowNet}~\cite{gu2019hplflownet}\\
\scriptsize \textbf{MeteorNet}~\cite{liu2019meteornet}\\
\scriptsize \textbf{PointPwcNet}~\cite{wu2019pointpwc}\\
\scriptsize \textbf{Flot}~\cite{puy20flot}, \textbf{EgoFlow}~\cite{tishchenko2020selfflow}
};

\node[color=stanfordgrey,anchor=south, align=left, execute at begin node=\setlength{\baselineskip}{2.1ex}] at (0.3,0.3) { \scriptsize \textbf{JGWTF}~\cite{mittal2019just}\\
\scriptsize \textbf{PointPwcNet}~\cite{wu2019pointpwc}\\
\scriptsize \textbf{EgoFlow}~\cite{tishchenko2020selfflow}
};

\node[color=stanfordgrey,anchor=south, align=left, execute at begin node=\setlength{\baselineskip}{2.2ex}] at (3.0,0.3) { \scriptsize \textbf{Ours}
};
\end{tikzpicture}
  }
  \vspace{-5mm}
  \caption{Recent scene flow methods either use full supervision (and suffer from domain gap) or no-supervision (and suffer from reduced performance). Instead, our method uses weak supervision and benefits from the best of both worlds.\vspace{-4mm}} %
  \label{fig:intro_supervision}
\end{figure}

Understanding dynamic 3D environments is a core challenge in computer vision and robotics. In particular, applications such as self driving and robot navigation rely upon a robust perception of dynamically changing 3D scenes. %
To equip autonomuous agents with the ability to infer spatiotemporal geometric properties, there has recently been an increased interest in \emph{3D scene flow} as a form of low-level dynamic scene representation~\cite{liu2019flownet3d,tishchenko2020selfflow,wang2019flownet3d++,puy20flot,niemeyer2019occupancy,rempe2020caspr}. 
Scene flow is the 3D motion field of points in the scene~\cite{vedula1999three} and is {a generalization} of 2D optical flow.
In fact, optical flow~\cite{beauchemin1995computation,horn1981determining} can be understood as the projection of the scene flow onto a camera image plane~\cite{devernay2006multi}. 
Such dense motion fields can serve bottom-up approaches for high-level dynamic scene understanding tasks like semantic segmentation~\cite{liu2019meteornet} or motion perception. However, representing dynamics via a free form velocity field has two major disadvantages. First, in most applications of interest, dynamics are attributed to rigid object motion~\cite{carceroni2002multi,menze2015joint}. %
\review{This notion has been extensively exploited in robotics~\cite{byravan2017se3,deng2020deep,bui20206d,birdal2018bayesian}} and holds especially for vehicles in autonomous driving. Predicting unconstrained per-point flow may lead to non-viable results, e.g. parts of the same car might move in different directions. Second, %
{accurately learning} direct flow estimation necessitates dense supervision that is expensive to acquire and prone to annotation errors. As a result, many methods have resorted to training on simulated data \cite{mayer2016FT3D,wang2019flownet3d++,puy20flot}, yet this comes at the price of a non-negligible domain gap. Other methods have attempted to solve the problem in a completely unsupervised manner \cite{tishchenko2020selfflow,wu2019pointpwc,mittal2019just}, however they fail to provide competitive performance. In~\cref{fig:intro_supervision} we illustrate these two extremes of no- and full-supervision while spanning the many intermediate possibilities, sorted according to annotation effort\footnote{Note that no prior work exists at different points on the spectrum given in~\cref{fig:intro_supervision}. This leaves ample room for future exploration.}. 
\insertimageC{1}{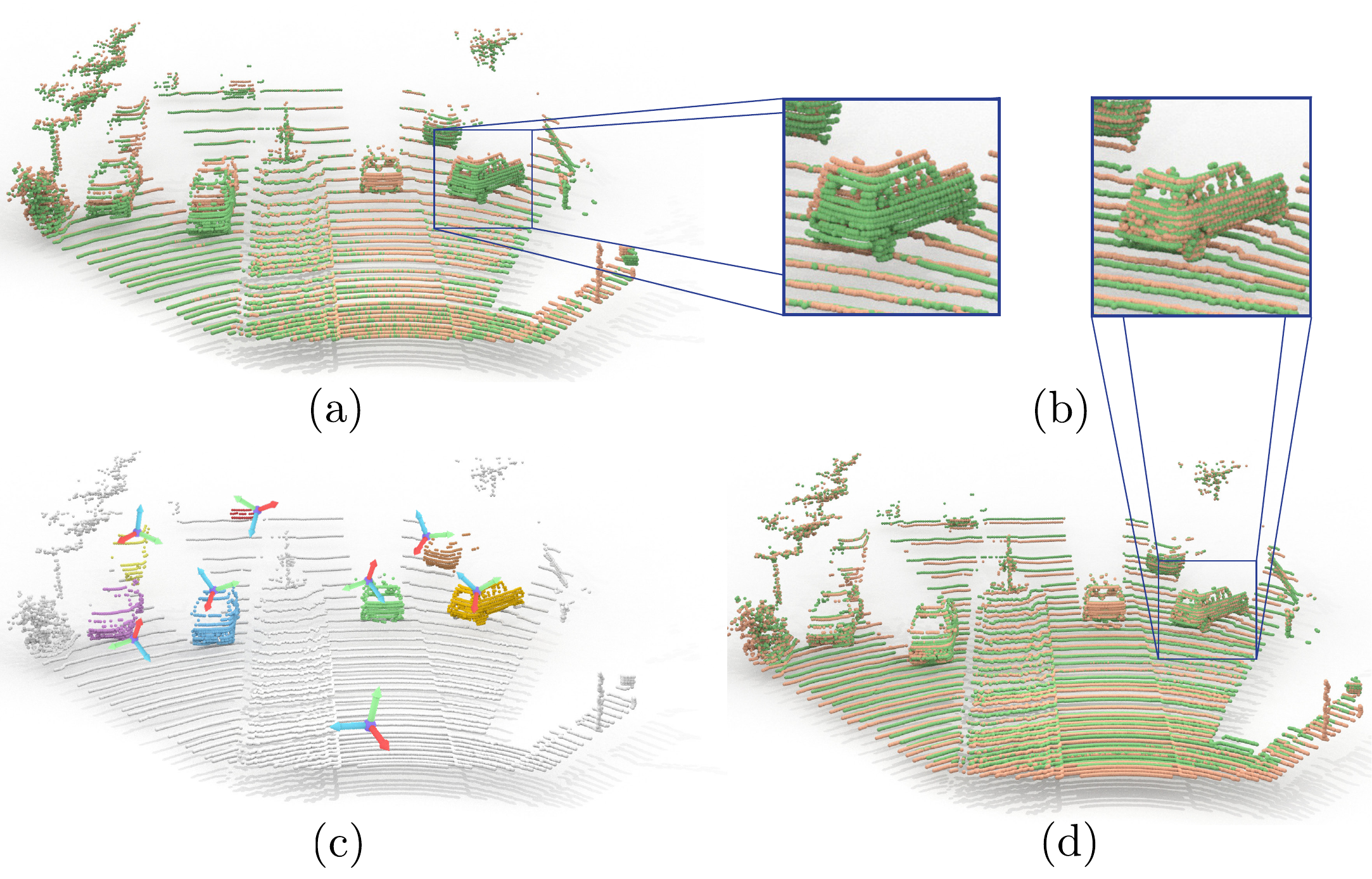}{Our network takes two successive frames as input (a), and outputs a set of transformation parameters for each segmented rigid agent (c) which are used to recover per-point rigid scene flow. After applying the predicted flow to the first point cloud, the two frames are aligned (b, d).\vspace{-4mm}}{fig:teaser_figure}{t!}

Based on this observation, %
we seek a sweet spot between %
supervision effort and performance. To this end, we propose a scene abstraction approach that uses rigid objects as the basic components. More specifically, by splitting the scene into foreground (movable objects) and background (static objects), we explain the background (BG) flow as the sensor ego-motion and the foreground (FG) flow as clusters of rigidly moving entities. As a result, we tackle both aforementioned challenges at the same time: (1) we enforce a rigidity constraint to get meaningful and more accurate (foreground and background) flow, (2) we can relax the requirement for dense flow supervision with a much simpler binary mask annotation and ego-motion that can often be extracted directly from the agent's IMU. The result is a weakly supervised method for accurate flow estimation that, unlike completely unsupervised approaches, outperforms previous state of the art (SoTA) by a significant margin. For example, reducing the end-point-error on the \emph{lidarKITTI} dataset by more than $30$~cm relative to the SoTA. At the same time our result provides an interpretable and readily usable object-level scene representation.  
In brief, our contributions are:
\begin{itemize}[noitemsep,topsep=3.5pt,leftmargin=\parindent]
 \item We exploit the geometry of the rigid scene flow problem to introduce an inductive bias into our network. This allows us to learn from weak supervision signals: background masks and ego-motion.
 \item Our data-driven method decomposes the scene into rigidly moving agents enabling us to reason on the level of objects rather than points. We use this notion to propose a new test-time optimization, further refining the flow predictions.
 \item Our method is backed by a novel, flexible, scene flow backbone, which can be adapted to solve various tasks.
\end{itemize}%
As a result of these contributions, our method greatly outperforms the SoTA on several benchmarks: \emph{FT3D}~\cite{mayer2016FT3D}, \emph{stereoKITTI}~\cite{menze2015joint}, \emph{lidarKITTI}~\cite{Geiger2012CVPR}, and \emph{semanticKITTI}~\cite{behley2019iccv}, while generalizing to the \emph{waymo-open} dataset~\cite{sun2019scalability} without additional fine-tuning. %

\insertimageStar{0.95}{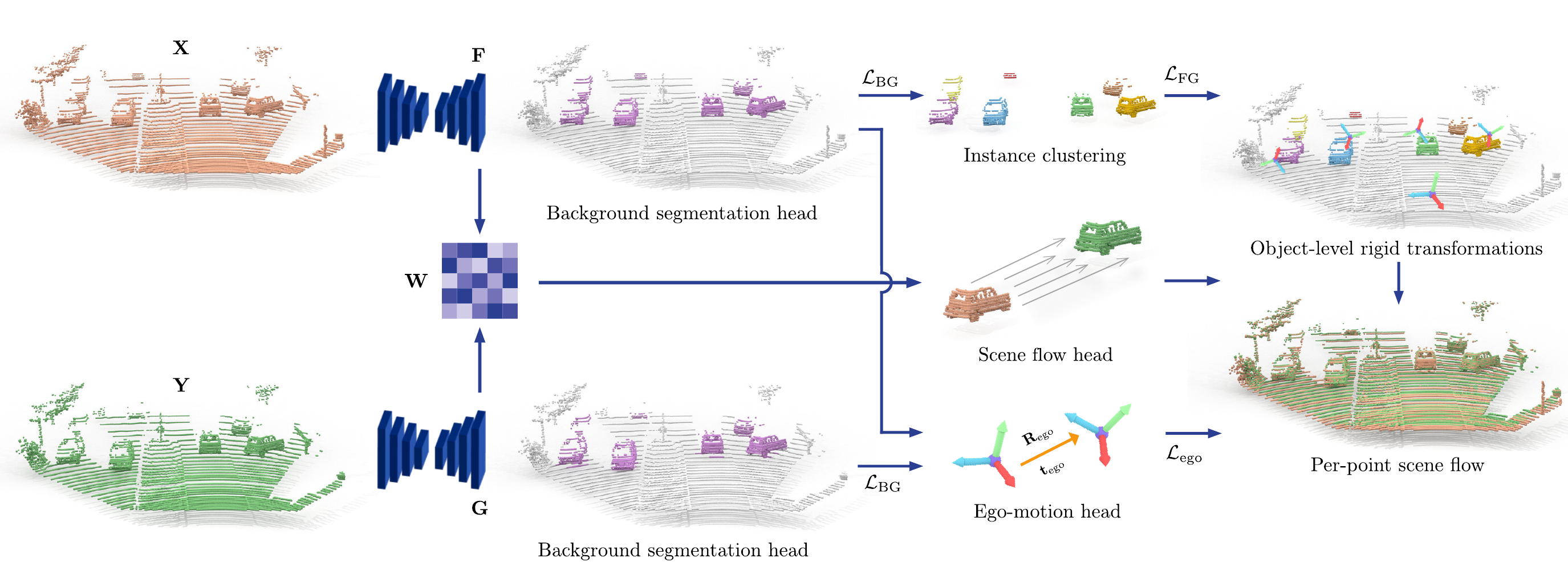}{Architecture of our weakly-supervised scene flow estimation pipeline. Our module consumes point clouds $\X$ and $\Y$ of two consecutive frames and estimates per-object transformation parameters $\{\mathbf{T}\}_{k=1}^{K-1}$, ego-motion $\T_{\mathrm{ego}}$, and object masks $\{\mathbf{z}\}_{k=1}^K$. These outputs can be combined into an object-level scene abstraction and pointwise rigid scene flow.
\vspace{-4mm}}{fig:SFpipeline}{t!}
\vspace{-1mm}
\section{Related Work}
\vspace{-2mm}
\label{sec:related}
\parahead{Data Driven 3D Scene Flow} 
While there is extensive literature on traditional 3D scene flow~\cite{vedula1999three,huguet2007variational,Wedel-et-al-08,jaimez15icra,tanzmeister2014grid,ushani2017learning,newcombe2015dynamicfusion,slavcheva2017killingfusion,slavcheva2018sobolevfusion,innmann2016volumedeform,dou2016fusion4d,devernay2006multi,carceroni2002multi,pons2005modelling}, we focus our attention to recent data-driven methods that emerged based on advances in deep learning on unordered point sets~\cite{qi2017pointnet,zaheer2018deep,ahmed2018deep}.

Early methods for 3D scene flow estimation mimicked their 2D counterparts. SceneFlowNet~\cite{mayer2016FT3D} used \textit{2D optical flow and disparity maps} to estimate the 3D scene flow.  FlowNet3D~\cite{liu2019flownet3d} successfully adopted the ideas from FlowNet~\cite{ilg2017flownet,dosovitskiy2015flownet}. 
FlowNet3D++~\cite{wang2019flownet3d++} extended FlowNet3D to incorporate additional geometric constraints.
MeteorNet~\cite{liu2019meteornet} used multiple temporally ordered frames to improve the accuracy of the inferred flow. 
Wang~\etal~\cite{wang2018deep} incorporated a continuous convolution into a 3D-FCN~\cite{tchapmi2017segcloud} to undo both the ego-motion and object-motion in two consecutive LiDAR frames. 
PointRNN \cite{fan2019pointrnn} used recurrent neural networks to model temporal point sets, which yields 3D scene flow as a by-product.
HPLFlowNet~\cite{gu2019hplflownet} ordered the points into a permutohedral lattice of SplatNet~\cite{su2018splatnet} to apply bilateral convolution layers. This allowed efficient and robust non-rigid 3D flow computation. 
Both OccFlow~\cite{niemeyer2019occupancy} and CaSPR~\cite{rempe2020caspr} introduced spatio-temporal representations to continuously and densely estimate the scene flow.
Mustafa and Hilton used semantic coherence between multiple frames to improve 4D scene flow estimation, co-segmentation and reconstruction~\cite{mustafa2019semantically}. 
Mittal~\etal~\cite{mittal2019just} and PointPWCNet~\cite{wu2019pointpwc} proposed self-supervised losses to infer the scene flow in an end-to-end manner. Finally, FLOT~\cite{puy20flot} proposed a simple correspondence-based end-to-end scene flow network.
\review{While our backbone also estimates correspondences, decomposing the scene into rigid agents provides us further higher level scene understanding and enables test-time optimization, while requiring less supervision.}

\parahead{Local Rigidity and Multi-body Motion} 
Flow estimation has also been tackled by imposing physical priors such as multi-body rigidity. Initial attempts involved factorization~\cite{costeira1998multibody,kanatani2001motion} to separate independently moving objects. 
Golyanik~\etal~\cite{golyanik2017multiframe} used rigidity constraints on over segmentation of RGBD-frames.
 \cite{moosmann2013joint} used detection \& tracking to constrain the flow. Vogel~\etal~\cite{vogel2013piecewise} modeled scene flow using piecewise rigidly moving planar patches. Dewan~\etal~\cite{dewan2016rigid} used 3D descriptors to enforce local geometric constancy on a factor graph.
GraphFlow~\cite{alhaija2015graphflow} and SphereFlow~\cite{hornacek2014sphereflow} considered large motions and relied on sparse keypoints that are not repeatable in 3D~\cite{salti2011performance,yang2018toward}. Similar to us, Jamiez~\etal~\cite{jaimez2017fast} as well as recent Dynamic-SLAM pipelines~\cite{huang2019clusterslam,saputra2018visual,strecke2019fusion} assumed that clustering would yield motion segmentation. In fact, MaskFusion~\cite{runz2018maskfusion} and EMFusion~\cite{strecke2019fusion} explicitly used Mask-RCNN~\cite{he2017mask} to this end. %

On a data driven front, considering the rigidity in the scenes~\cite{lv2018learning}, Ma~\etal~\cite{ma2019deep} made use of depth and flow estimates from a stereo RGB-D setup within an optimization framework to obtain the 3D motion of each instance. This method relied upon a given instance segmentation, a difficult problem to solve even for the SoTA approaches~\cite{wang2018sgpn,yi2019gspn,yang2019modeling}. 
Based on VoxelNet~\cite{zhou2018voxelnet}, PointFlowNet~\cite{behl2019pointflownet} jointly predicted 3D scene flow, bounding boxes, and rigid motion of objects in the scene. 
Yi~\etal~\cite{yi2019deep} used a PointNet++~\cite{qi2017pointnet++} based flow estimator for piecewise rigid 3D part induction.

\vspace{-8mm}
\section{Method}
\label{sec:method}
\vspace{-2mm}
\parahead{Problem setting}
Suppose that we observe a pair of 3D scenes $\X$ and $\Y$ acquired by a \emph{single moving observer} in two consecutive instants $t_0$ and $t_1$, respectively. Here, $\X\in\R^{3\times N} = \{\x_i \in \R^3\}_i$ denotes a point cloud (so does $\Y$) and $\V\in\R^{3\times N} = \{\v_i \in \R^3\}_i$ its corresponding vector field in 3D s.t. $\X+\V\apeq\Y$ relates frame $t_0$ to $t_1$. We further assume that $\X$, $\Y$, and also $\V$ are \emph{multi-body} \ie composed of multiple objects. Hence, $\V$ can be clustered into $K$ objects $\flow=\{\V_k\in\R^{3\times N_k}\}_{k=1}^{K}$ each of which follows rigid dynamics, \ie $\V$ can be summarized by a set of $K$ rigid transformations $\Tset\equiv\{\T_k\in SE(3)\}_{k=1}^K$ such that $\V\apeq\{\T_k\circ{\X}_k-{\X}_k\}_{k=1}^K$ where $K \ll N$ and:
\begin{equation}
\label{eq:se3}
SE(3) = \left \{
\T \in \R^{4\times 4} \colon \T=\begin{bmatrix} 
\Rot & \tra \\ 
\zero^\top & 1 
\end{bmatrix}
\right\},
\end{equation}
$\Rot \in SO(3),\, \tra \in \R^3$.\footnote{We denote the action of $\T$ as ${\X}^\prime = \T\circ\X$ and $\hat{\X}^\prime= \T\hat{\X}$ where $\hat{\X}\in\R^{4\times N}$ is the \emph{homogenized} $\X$.} %
The motion of the immobile \emph{background}, determines the \emph{ego-motion} $\Tego\subset \Tset$.

\parahead{Summary}
We refrain from directly predicting unconstrained pointwise flow vectors and rather aim to estimate $\Tset\equiv\{\T_k\}_{k=1}^K$ for all rigid bodies from which the entire scene flow $\flow=\{\V_k\}_{k=1}^K$ can be recovered. To this end, we propose to \emph{learn} the task of rigid flow estimation by solving an optimization problem composed of a set of loss functions as illustrated in~\cref{fig:SFpipeline}. To reason on the level of objects, we use both instance masking and motion. To obtain the masks, we use a FG / BG prediction module in conjunction with an FG clustering. To estimate ego-motion, we run a differentiable registration on the BGs extracted from both point sets. The motions of the individual objects in the FGs are obtained similarly under the assumption of local-rigidity. We avoid flow-level or instance-level supervision altogether and only assume the availability of the \emph{binary FG/BG annotations} and \emph{ego-motion information} as a much weaker supervision signal than dense scene flow.
In the sequel, we first describe our formulation of the individual objectives (\cref{sec:energy}) before proceeding to our network architecture (\cref{sec:inference}). Additional details are available in our supplement. %

\subsection{Energy Formulation}\label{sec:energy}
Our solution to the 3D scene flow estimation is attained as the minimum of a non-convex energy composed of a BG segmentation loss $\loss_{\mathrm{BG}}$, an ego-motion loss $\loss_{\mathrm{ego}}$, and an FG loss $\loss_{\mathrm{FG}}$: %
\begin{align}\label{eq:loss_functions}
     \pars^\star &=\argmin_{\pars} \loss_{\mathrm{BG}} + \loss_{\mathrm{ego}} + \loss_{\mathrm{FG}} %
\end{align}
where the optimal rigid scene flow $(\flow^\star, \Tset^\star)$ results from the output of a \emph{deep neural network} $\varphi$ with learnable parameters $\pars$: $(\flow^\star, \Tset^\star) = \varphi_{\pars^\star}(\X, \Y)$. Next, we detail the individual loss terms; each involves an \emph{unknown} or \emph{latent variable} obtained as a network prediction as specified in~\cref{sec:inference}. 

\parahead{Background segmentation error ($\loss_{\mathrm{BG}}$)}
\label{sec:method_segmentation}
To decompose the scene into agents that move as rigid bodies, we follow a coarse-to-fine approach. In the first step we aim to split the background and foreground points, where the foreground represents all points belonging to the \review{movable} objects (e.g., cars, cyclists, people, \dots). In order to learn this binary segmentation of a \review{point cloud}, 
we minimize the loss $\loss_{\mathrm{BG}} = \frac{1}{2}(\loss_{\mathrm{BG}}^\X + \loss_{\mathrm{BG}}^\Y)$ where: %
\begin{equation}
\loss^\X_{\mathrm{BG}} = \frac{1}{N} \sum_{i=1}^{N} \mathrm{BCE}(h_i^\X, \bar{h}_i^\X).
\label{eq:loss_BG}
\end{equation} 
$\hgt^\X$, $\hgt^\Y$ denote the GT binary masks of point clouds $\X$ and $\Y$, respectively. $\h^\X=\{h_i^\X\}_{i=1}^N$ and $\h^\Y=\{h_i^\Y\}_{i=1}^N$ are the \review{inferred} foreground probabilities of points in $\X$ and $\Y$ yet to be clarified in~\cref{sec:inference}, and
$\mathrm{BCE}(h_i, \bar{h}_i) = \overbar{h}_i\log(h_i) + (1 - \overbar{h}_i)\log(1 - h_i)$.

\parahead{Ego-motion error ($\loss_{\mathrm{ego}}$)}
\label{sec:method_ego}
The scene flow of the physically static background can be fully explained by its transformation parameters, the ego-motion. We estimate these parameters by first extracting the background points of both the source and target point cloud\footnote{During training we use the GT BG segmentation mask $1 - \hgt$, while during inference we threshold the inferred FG probabilities $1-{\h}$.}. \review{To reduce the computational complexity}, we then randomly sample $N^b=1024$ points therefrom such that $\X^b \in \R^{3 \times N^b} \subset \X$ and $\Y^b \in \R^{3 \times N^b} \subset \Y$. The goal of ego-motion estimation is to compute the optimal $\Rego \in \SOg$ and $\tego \in \R^3$ in the weighted least-squares sense
\begin{equation}
    \Rego^\star, \tego^\star = \argmin_{\Rego,\tego} \sum_{l=1}^{N^b}  w_l \| \Rego\x^b_l + \tego - \phi({\x^b_l,\Y^b}) \|^2, 
    \label{eq:weighted_pairwise}\nonumber
\end{equation}
where $\phi(\x,\Y)$ is a \emph{soft assignment function} returning a point from $\Y$ that corresponds to $\x\in\X$. The weights $w_l$ will be clarified below. %

We approximate the optimal soft assignment $\phi$ via the entropy-regularized Sinkhorn algorithm~\cite{sinkhorn1964relationship,cuturi2013sinkhorn,yew2020rpm}. To this end, we momentarily assume the availability of an affinity matrix $\M \in  \R^{N^b \times N^b}$ describing the similarity of the background points in  $\X^b$ and $\Y^b$. Prediction of the currently unknown $\M$ will be made precise in~\cref{sec:inference}.
Given $\M$, we perform an alternating row and column normalization on it for $k_S\triangleq 3$ iterations, which yields $\A \in \R_{+}^{N^b \times N^b}$, a \emph{doubly stochastic} (DS) assignment matrix. 

In practice, due to the occlusions and sampling pattern, not all background points will have correspondences. We therefore add a slack row and column to $\M$ (hence to $\A$), which enable down-weighting the outliers, while still returning a DS-matrix. The soft correspondence function then reads $\phi({\x_i^b,\Y^b})=
{\Y^b\mathbf{a}_i}/{\| \mathbf{a}_i\|_1}$ where $\mathbf{a}_i$ is the $i$-th column of $\A$ after removing the slack row. %
\review{Given the correspondences,} the ego-motion can be recovered \review{in closed-form} using a (differentiable) \emph{weighted} Kabsch algorithm~\cite{kabsch1976solution} \review{where the weights $w_i$ are obtained as the total contribution $w_i = \sum_{j=1}^{N^b} a_{ij}$.}  
\review{Our ego-motion penalty measures the $l1$-discrepancy between the points transformed with the estimated $(\Rego,\tego)$ and the GT parameters $(\overbar{\Rot}_\mathrm{ego},\overbar{\tra}_\mathrm{ego})$:}
\begin{equation}
    \loss_\mathrm{trans} = \frac{1}{B} \sum_{i=1}^B \|(\overbar{\Rot}_\mathrm{ego}\x^b_i + \overbar{\mathbf{t}}_\mathrm{ego}) - (\Rego\x^b_i + \tego)\|_1,\nonumber
\end{equation}
where $B$ denotes the number of all background points in point cloud $\X$. To stabilize the training, we further add a regularizer loss that discourages the assignment of large values to the slack rows and columns~\cite{yew2020rpm}:
\begin{equation}
    \loss_\mathrm{inlier} =  \frac{1}{N^b} \sum_{i=1}^{N^b}\Big(1- \sum_{j=1}^{N^b} a_{ij}\Big) + \frac{1}{N^b} \sum_{j=1}^{N^b}\Big(1- \sum_{i=1}^{N^b} a_{ij}\Big).\nonumber
\end{equation}
The overall ego-motion loss is then the weighted sum: %
\begin{align}
    \loss_{\mathrm{ego}} = \loss_\mathrm{trans} + \lambda_\mathrm{inlier}\loss_\mathrm{inlier}
\end{align}
where $\lambda_\mathrm{inlier} := 0.005$ in all our experiments.

\parahead{FG instance-level rigidity error ($\loss_{\mathrm{FG}}$)}
\label{sec:object_rigid}
To define our \emph{per-instance rigidity loss}, we assume the availability of the following entities: (\textbf{i}) $\X^f$, the points of the source frame belonging to the FG; (\textbf{ii}) $\V^f$, the flow vectors associated to $\X^f$; and (\textbf{iii}) foreground clusters $\clus = \{\C^k\in\R^{3\times N_k}=\{\cx^k_j\in\R^3 \}_j\}^{N^C}_{k=1}$ aggregating the individual rigid entities. (i) is a by-product of BG segmentation, \ie during training, the indices of these points are obtained from the GT mask and during inference by thresholding the \emph{inferred} FG probabilities. The flow vectors in (ii) are the result of the \emph{scene flow module} (see~\cref{sec:inference}). Finally, (iii) is computed by a simple DBSCAN clustering~\cite{ester1996density} of the 3D coordinates in $\X^f$, which is based on the hypothesis that the foreground objects scattered across the scene are naturally separated by \emph{void space}~\cite{jiang2020pointgroup}. We refrain from using a data driven instance segmentation module, because our simple approach alleviates the need for instance segmentation labels. 

Our rigidity loss $\mathcal{L}_\mathrm{rigid}$ encourages the predicted flow vectors $\V^f_k$ of each cluster $k$ to be \emph{congruent}, \ie $\V^f_k$ can be well approximated by a rigid transformation $\T_k$ composed of the rotation $\Rot_k$ and translation $\tra_k$: 
\begin{equation}
    \loss_{\mathrm{rigid}} = \frac{1}{N^c} \sum^{N^c}_{k=1} \frac{1}{N^k} \sum^{N^k}_{j=1} \|\Rot_k\cx^k_j + \tra_k - (\cx^k_j + \v^k_j)\|_1
\end{equation} %
\review{The supervision signals} $\Rot_k$ and $\tra_k$ are computed on the fly, such that they best explain the underlying flow $\V^f$:
\begin{equation}\label{eq:TransV}
\T_k^\star = \argmin_{\T_k} \|\T_k\circ \C_k - (\C_k + \V_k^f)\|.
\end{equation}
We solve~\cref{eq:TransV} for each individual cluster once again using the Kabsch algorithm~\cite{kabsch1976solution}. 
We additionally complement the per-cluster rigidity objective with a two way Chamfer distance (CD) computed across all the foreground points:
\begin{equation}
\loss_\mathrm{CD} = \sum_{\x \in \X^f_v }  \min_{\y \in \Y^f} \|\x - \y\|_2 + \sum_{\y \in \Y^f}  \min_{\x \in \X^f_v} \|\x - \y\|_2
\end{equation}
where $\X^f_v := \X^f + \V^f$. The overall \review{FG} loss is then a weighted sum of the above objectives:
\begin{equation}
    \loss_{\mathrm{FG}} = \loss_\mathrm{rigid} + \lambda_\mathrm{CD}\loss_\mathrm{CD}
\end{equation}
where $\lambda_\mathrm{CD} := 0.5$ in all our experiments.

\subsection{\review{Network implementation}}\label{sec:inference}
Provided a large dataset with \emph{FG-BG mask} annotations as well as \emph{ego-motion}, we learn to minimize~\cref{eq:loss_functions}
using a deep neural network $\varphi_{\pars}$ as shown in~\cref{fig:SFpipeline}. In the sequel, we describe the individual modules of our network and the full inference of rigid scene flow. We refer the reader to the supplement for more details. 

\parahead{Backbone}\label{sec:backbone}
Our formulation (\cref{sec:energy}) involves the estimation of different entities, requiring our network to solve \emph{multiple tasks} similar to~\cite{jiang2019sense}. While it would be possible to deploy a specialized network for each task, this would increase the memory footprint and would not encourage tasks to reinforce each other~\cite{zamir2018taskonomy,zamir2020robust}. Instead, we propose a flexible backbone suited for solving multiple tasks through specialized heads. 
Specifically, our backbone network is based on Minkowski-Net~\cite{choy20194d} and follows a U-Net~\cite{ronneberger2015u}-like encoder-decoder architecture with skip connections. Its input is a \emph{sparsely} voxelized point cloud $\X^v \in \R^{3 \times N^v }$ and its outputs are per-point latent features $\F^v \in \R^{64 \times N^v}$. The same backbone with shared weights is also applied to $\Y^v \in \R^{3 \times M^v}$ to obtain the latent features $\G^v \in \R^{64 \times M^v}$. In the following, we omit the superscript $v$ for clarity and unless specified differently, use $\X$ and $\Y$ to refer to the voxelized point clouds and $\F$ and $\G$ to refer to their associated latent features.

\parahead{Background segmentation head}
Our background segmentation head consists of two sparse convolutional layers with instance normalization and the ReLU~\cite{nair2010rectified} activation function after the first one. It takes the latent features $\F$ and $\G$ as input and outputs per-point foreground probabilities $\mathbf{h^\X} \in \R^{N^v}$ and $\mathbf{h^\Y} \in \R^{M^v}$.

\parahead{Ego-motion head}
Given the latent features  $\F^b$ and $\G^b$ of the background points $\X^b$ and $\Y^b$, the ego-motion head computes the affinity matrix $\M \in  \R^{N^b \times N^b}$ s.t.
\begin{equation}
M_{ij} = \exp\big(-{ \|\mathbf{f}^b_i - \mathbf{g}^b_j\| }/{\tau_\mathrm{ego}}\big),
\end{equation}
where  $\tau_{\mathrm{ego}}$ 
controls the \emph{softness} of the correspondences.

\parahead{Scene flow head}
\label{sec:method_flow}
Based on the notion that scene flow is tightly coupled to correspondences~\cite{puy20flot}, we now devise our scene flow head. To assure differentiability, we estimate \emph{soft} correspondences. To allow for large motion, we measure the similarity in the latent space of \emph{features} $\F$ and $\G$ instead of in the physical space. Hence, we find the corresponding points in $\X$ and $\Y$ as: 
\begin{equation}
    \X_c := \Y \D , \,\,\, d_{ij} := \mathrm{softmax}(-\frac{1}{\tau_\mathrm{flow}}\|\mathbf{f}_i - \mathbf{g}_j\|_2)
\end{equation}
where $\tau_\mathrm{flow}$ is again a learnable temperature that controls the \emph{softness} of the correspondences in the same manner as $\tau_\mathrm{ego}$ above. 
Soft correspondences can be used to compute the initial flow estimate as $\V^{\mathrm{init}} :=\X_c - \X$. However, this initial flow vector field is likely to be noisy due to large motions, sampling, and imperfect latent features and thus still has to be refined~\cite{puy20flot}.
Our \emph{refinement} module takes $\V^\mathrm{init}$ as input and locally smoothens it by estimating a residual flow $\Delta\V^\mathrm{init}$ thorough a series of sparse convolutional layers. The refined scene flow $\V \in \R^{3 \times N^v}$ is then obtained as: $\V = \V^\mathrm{init} + \Delta\V^\mathrm{init}$. The detailed architecture of the scene flow head is given in the supplement.

\parahead{From transformations to per-point rigid scene flow}
The output of our multi-task network comprises of: (i) transformation parameters of the ego-motion $\Tego$ and individual clusters $\{\T_k\}_{k=1}^{K-1}$; (ii) object level masks $\{\mathbf{z
}_k\}_{k=1}^{K}$; and (iii) unconstrained pointwise scene flow estimates $\V$. \review{The \emph{pointwise rigid scene flow} $\V^\mathrm{rigid}$ can then be recovered as $\V^\mathrm{rigid} \apeq \{\T_k\circ {\X}_k-{\X}_k\}_{k=1}^K$, where $\X_k$ denotes the points of $\X$ belonging to the cluster $k$ according to the inferred object masks $\mathbf{z}_k$. For the points that are neither assigned to the background nor to any of the foreground rigid bodies, we use the unconstrained scene flow predictions $\V$.}  

\parahead{Training and implementation details}
Our method is implemented in PyTorch using the \emph{MinkowskiEngine}~\cite{choy20194d}. Unless specified differently, we train our network in an end-to-end manner for $40$ epochs (or until convergence), by minimizing \cref{eq:loss_functions}. We train on a single NVIDIA GTX2080Ti with batch size $8$. We use the Adam~\cite{kingma2014adam} optimizer with an initial learning rate $10^{-3}$, which is decayed every epoch according to an exponential schedule with $\gamma = 0.98$. The whole training takes about two and a half days. \review{The detailed parameters of the Sinkhorn algorithm and DBSCAN clustering are available in the supplement.}

\parahead{\review{Inference}} \review{The abstraction of the scene into a collection of rigid bodies enables us to run test-time optimization of their inferred transformation parameters. Specifically, we run an optimization scheme in which we iteratively minimize the closest point distance to the target points. For ego-motion we index the background points, and for individual clusters, all the foreground points of $\Y$. The indexing is performed using the inferred object masks $\mathbf{z}_k^\mathbf{X}$ and $\mathbf{z}_k^\mathbf{Y}$. Given the final transformation estimates $\{\T^\star_k\}_{k=1}^{K}$ \emph{pointwise rigid scene flow} can be recomputed as $\V^\mathrm{rigid} \apeq \{\T^\star_k\circ {\X}_k-{\X}_k\}_{k=1}^K$. A detailed description of our optimization scheme including its run-time is available in the supplement.}

In the following, we reintroduce the superscript $v$ so as to denote the voxelized point $\X^v$. Note that $\V^\mathrm{rigid}$ represents the flow for the voxel centers $\X^v$ and during inference, still has to be \emph{transferred} to the points $\X$. We perform this transfer by a simple inverse-distance weighted interpolation:
\begin{equation}
    \v^\star_i = \frac{\sum_{j:\x^v_j \in \Edge(\x_i)} \mathbf{v}^\mathrm{rigid}_j \|\x_i - \x^v_j\|_2\inv}{\sum_{j:\x^v_j \in \Edge(\x_i)} \|\x_i - \x^v_j\|_2\inv},
    \label{eq:voxel_to_points}
\end{equation}
where $\Edge(\cdot)$ returns the set of $k$-NN in the Euclidean sense.

\section{Experimental Evaluation}
\label{sec:experiments}
In this section, we first describe the datasets (\cref{sec:experiments_datasets}) and evaluation metrics (\cref{sec:experiments_metrics}) used in our experiments. \review{We start the evaluation, by assessing the performance of} our backbone flow estimation network under full supervision on point clouds lifted from stereo images~(\cref{sec:experiments_backbone}). We \review{then} proceed to evaluate our full pipeline in a weakly supervised setting on real LiDAR scans (\cref{sec:experiments_weakly_supervised}). Finally, we showcase the generalization capability of our method (\cref{sec:generalization}) and justify our design choices in an ablation study (\cref{sec:ablation}).
\vspace{-0.2em}
\subsection{Datasets}
\begin{table}[!t]
    \setlength{\tabcolsep}{4pt}
    \renewcommand{\arraystretch}{1.25}
	\centering
	\resizebox{\columnwidth}{!}{
    \begin{tabular}{clccccc}
            \toprule
			Dataset & Method & Supervision & EPE3D [m]~$\downarrow$ & Acc3DS~$\uparrow$ & Acc3DR~$\uparrow$ & Outliers~$\downarrow$ \\
            \hline
            \multirow{6}{*}{\emph{FT3D}} & FlowNet3D~\cite{liu2019flownet3d} & Full & 0.114 & 0.412 & 0.771 & 0.602 \\
            & HPLFlowNet~\cite{gu2019hplflownet} & Full & 0.080 & 0.614 &0.855 & 0.429 \\
			& PointPWC-Net~\cite{wu2019pointpwc} & Full & 0.059 & 0.738 & 0.928 & \textbf{0.342} \\
			& FLOT~\cite{puy20flot} & Full & \textbf{0.052} & 0.732 &0.927 & 0.357 \\
			& EgoFlow~\cite{tishchenko2020selfflow} & Full & 0.069 & 0.670 &0.879 & 0.404 \\
			& Ours & Full & \textbf{0.052} & \textbf{0.746} & \textbf{0.936} & 0.361 \\
			\hline
			\multirow{6}{*}{\emph{stereoKITTI}}
			& Flownet3D~\cite{liu2019flownet3d} & Full & 0.177 & 0.374 & 0.668 & 0.527 \\
            & HPLFlowNet~\cite{gu2019hplflownet} & Full & 0.117 & 0.478 & 0.778 & 0.410 \\
			& PointPWC-Net~\cite{wu2019pointpwc} & Full & 0.069 & 0.728 & 0.888 & 0.265 \\
			& FLOT~\cite{puy20flot} & Full & 0.056 & 0.755 & 0.908 & 0.242 \\
			& EgoFlow~\cite{tishchenko2020selfflow} & Full & 0.103 & 0.488 &0.822 & 0.394 \\
			& Ours & Full & \textbf{0.042} & \textbf{0.849} & \textbf{0.959} & \textbf{0.208} \\
			\bottomrule
	\end{tabular}
	}
	\vspace{0.1mm}
	\caption{Evaluation results in a fully supervised setting on \emph{FT3D} and \emph{stereoKITTI} datasets.}
	\vspace{-0.3cm}
	\label{tab:flow_traditional}
\end{table}
\insertimageStar{0.945}{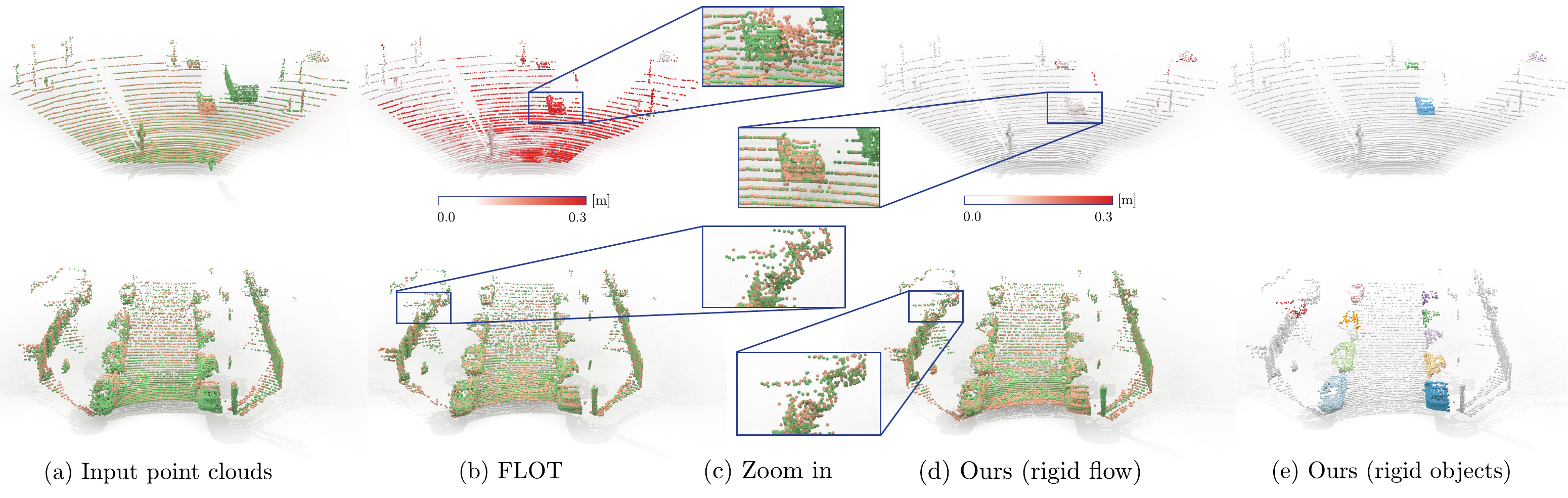}{Qualitative results of our weakly supervised method on \emph{lidarKITTI} (top) and \emph{waymo open} (bottom). For improved visibility, the \emph{EPE3D} (top row b,c ) is clipped to the range between 0.0~m (white) at 0.3m (red). As a result of predicting an unconstrained pointwise sceneflow, the rigid objects (car) in the results of FLOT might get deformed (d).\vspace{-1.3em}
}{fig:qualitative_results}{t!}
\label{sec:experiments_datasets}
For all datasets, we follow a common preprocessing step~\cite{liu2019flownet3d,gu2019hplflownet} and remove points whose depth or distance to the sensor is larger than $35$~m. For training and evaluation, we randomly sample 8192 points from both frames independently. A detailed description of the datasets and preprocessing steps is available in the supplement.

\parahead{FlyingThings3D (FT3D)}~\cite{mayer2016FT3D} is a large-scale stereo dataset of synthetic man-made objects that are scattered in space and move randomly between the two frames. We generate the point clouds and GT scene flow in accordance with~\cite{gu2019hplflownet}. FT3D consists of 19640 training examples (from which we use 3928 for validation) and 3824 test examples. Note, FT3D is only used for training in the fully supervised evaluation of our backbone and scene flow head (\cref{sec:experiments_backbone}).

\parahead{stereoKITTI}~\cite{menze2015joint, menze2018object} is a real world scene flow dataset with 142 point cloud pairs, which are all used for testing. The point clouds and GT scene flow are obtained by lifting the annotated disparity maps and optical flow to 3D~\cite{gu2019hplflownet}. As a consequence, the points of the two frames are under direct correspondence. We remove the ground points by a naive thresholding of the height coordinate~\cite{liu2019flownet3d, gu2019hplflownet}.

\parahead{lidarKITTI}~\cite{Geiger2012CVPR} is a real world dataset acquired with a Velodyne 64-beam LiDAR. It consists of the same 142 pairs as \emph{stereoKITTI}. GT is obtained by projecting the point clouds to the image plane and assigning them the annotated 3D flow vectors. In this dataset, the points of the two input frames are not in direct correspondence and have a typical sampling pattern of a LiDAR sensor.

\parahead{semanticKITTI}~\cite{behley2019iccv} provides per point semantic labels and accurate ego-motion for 21 LiDAR sequences of the KITTI odometry dataset. It is split into eleven (00-10) LiDAR sequences for training and eleven (11-21) for testing. We use sequences 03 and 05 for validation and the remaining nine for training. SemanticKITTI is used to train our method in a weakly supervised manner (\cref{sec:experiments_weakly_supervised} to \cref{sec:ablation}). Note that this dataset does not contain dense scene flow annotations.
\vspace{-0.1cm}
\subsection{Evaluation metrics}
\vspace{-0.1cm}
\label{sec:experiments_metrics}
\begin{table}[!t]
    \setlength{\tabcolsep}{4pt}
    \renewcommand{\arraystretch}{1.15}
	\centering
	\resizebox{\columnwidth}{!}{
	\begin{threeparttable}[t]
    \begin{tabular}{ccccccc}
            \toprule
			Dataset & Method & Supervision & EPE3D [m]~$\downarrow$ & Acc3DS~$\uparrow$ & Acc3DR~$\uparrow$ & Outliers~$\downarrow$ \\
            \hline
            \multirow{7}{*}{\begin{tabular}{c}\emph{lidarKITTI}\\ (w/o ground) \end{tabular}} 
			& PointPWC-Net~\cite{wu2019pointpwc} & Full & 0.390 & 0.387 & 0.550 & 0.653 \\
			& FLOT~\cite{puy20flot} & Full & 0.653 & 0.155 & 0.313 & 0.837 \\
			& Ours (backbone) & Full & 0.535 & 0.262 & 0.437 & 0.742 \\
			& Ours & Weak & 0.150 & 0.521 & 0.744  & 0.450 \\ 
			& Ours+ & Weak & 0.110  & 0.745  & 0.844 & 0.353  \\
			& Ours++ & Weak & \textbf{0.094} & \textbf{0.784} & \textbf{0.885} & \textbf{0.314} \\
			\hline
			\multirow{8}{*}{\begin{tabular}{c}\emph{lidarKITTI}\\ (with ground) \end{tabular}} 
			& PointPWC-Net~\cite{wu2019pointpwc} & Full & 0.710 & 0.114 & 0.219 & 0.932 \\
			& FLOT~\cite{puy20flot} & Full & 0.773 & 0.084 & 0.177 & 0.943 \\
			& MeteorNet~\cite{liu2019meteornet}~\tnote{1} & Full & 0.277 & / & / & / \\
			& Ours (backbone) & Full & 0.820 & 0.102  & 0.190 & 0.934 \\
			& Ours & Weak &  0.133 &  0.460 & 0.746 & 0.527 \\
			& Ours+ & Weak & 0.106 & 0.673 &  0.808 & 0.421 \\
			& Ours++ & Weak & \textbf{0.102}  & \textbf{0.686}  & \textbf{0.819} & \textbf{0.410} \\
			\bottomrule
	\end{tabular}
	\begin{tablenotes}
     \item[1] MeteorNet uses three tine frames and was trained on 100 and evaluated on 42 scenes of this dataset.
   \end{tablenotes}
\end{threeparttable}
	}

	\caption{Evaluation results on \emph{lidarKITTI}. Ours (backbone) denotes our model from \cref{sec:experiments_backbone} trained with full supervision on \emph{FT3D}. Ours are the direct estimates of our pipeline. Ours+ and Ours++ additionally denote test-time optimization of only ego-motion and all rigid bodies, respectively.}
 	 \vspace{-1em}
	\label{tab:kitti_lidar_flow}
\end{table}

We use standard evaluation metrics to assess the performance of our approach and compare it with SoTA methods, \emph{FlowNet3D}~\cite{liu2019flownet3d}, \emph{HPLFlowNet}~\cite{gu2019hplflownet}, \emph{PointPWCNet}~\cite{wu2019pointpwc}, FLOT~\cite{puy20flot}, and EgoFlow~\cite{tishchenko2020selfflow}.
Our main evaluation metric is the 3D end-point-error (\emph{EPE3D}), defined as the mean $l2$ distance between the predicted and GT scene flow. 

Additionally, we follow~\cite{liu2019flownet3d, gu2019hplflownet} and also report: (i) strict accuracy (\emph{Acc3DS}), defined as the percentage of points whose \emph{EPE3D}~$< 0.05$~m or relative error $<0.05$, (ii) relaxed accuracy (\emph{Acc3DR}) that denotes the ratio of points whose \emph{EPE3D}~$< 0.10$~m or relative error $<0.10$, and (iii) \emph{Outliers}, i.e. the ratio of points whose \emph{EPE3D}~$> 0.30$~m or relative error $>0.10$.

For the experiments in a weakly supervised setting, we also report the relative angular error (\emph{RAE}) and the relative translation error (\emph{RTE}) of the estimated ego-motion.

\vspace{-0.1cm}
\subsection{Our backbone under full supervision}
\label{sec:experiments_backbone}
\vspace{-0.15cm}
A core module of our proposed pipeline is the backbone network described in \cref{sec:backbone}.
It is therefore valuable to first assess its performance in conjunction with the scene flow prediction head, before turning to evaluate the performance of our entire weakly-supervised pipeline. To this end, we follow the traditional setting used by our competitors and train in a fully supervised manner on \emph{FT3D} by minimizing the $l1$ distance between the predicted and GT scene flow. We then evaluate our model on both \emph{FT3D} and \emph{stereoKITTI}. 

When evaluated on \emph{FT3D}, our method performs on par with FLOT~\cite{puy20flot} in terms of \emph{EPE3D} and outperforms all methods in terms of \emph{Acc3DS} and \emph{Acc3DR} (\cref{tab:flow_traditional}).
More importantly, it achieves superior generalization performance on \emph{stereoKITTI}, where it consistently outperforms SoTA in all evaluation metrics, setting a new SoTA with $0.042$~m \emph{EPE3D} ($\approx 1.5$~cm better than the closest competitor).
Based on these results, we conclude that sparse convolutions are an effective backbone for scene-flow estimation, and that our simple scene flow head can match the performance of SoTA, while enabling better generalization.

\vspace{-0.1cm}
\subsection{Our pipeline under weak supervision}
\label{sec:experiments_weakly_supervised}
\vspace{-0.1cm}
\parahead{Setting} Point clouds in both \textit{FT3D} and \textit{stereoKITTI} are obtained in the same manner: by lifting stereo images to 3D. Hence, their domain gap is relatively small. On the other hand, in LiDAR-based autonomous driving scenarios, point clouds are much sparser and assume a very different sampling pattern, resulting in a much more challenging setting for scene flow estimation.

We evaluate our entire weakly-supervised pipeline in this challenging setting by using the \emph{lidarKITTI} dataset. Specifically, we consider two scenarios: 1) we remove ground points by naively thresholding the vertical coordinate, 2) we use the ``raw" point clouds that also include the ground points for which the flow estimation is especially difficult.

We train a joint model for both scenarios in a weakly supervised manner using the point clouds from \emph{semanticKITTI}. Unlike \emph{semanticKITTI}, \emph{lidarKITTI} only includes the points and annotations of the objects that are visible within the front camera images. We therefore process semanticKITTI in the same manner. Since there are no LiDAR datasets available with scene flow annotations, we use the models trained on \textit{FT3D} for all the baselines.

\parahead{Evaluation}
\cref{fig:qualitative_results} and \cref{tab:kitti_lidar_flow} show that the domain gap between the stereo and LiDAR point clouds is too big for the traditional fully supervised methods to generalize effectively. Indeed, the performance of SoTA methods is up to 10 times worse when compared to the results on \emph{stereoKITTI}. On the other hand, our weakly supervised model predicts accurate rigid scene flow with an \emph{EPE3D}~$\approx$~0.1~m for both scenarios (after test-time optimization), while also providing an object-level abstraction (\cref{fig:qualitative_results}). Since our fully supervised backbone model also fails to generalize, we conclude that the crucial advantage of our method does not lie in a stronger backbone, but rather in the ability to train on the same domain.
Additional qualitative and quantitative results are available in the supplement.

\vspace{-0.15cm}
\subsection{Generalization to other datasets.}
\label{sec:generalization}
\vspace{-0.1cm}
\emph{Waymo open}~\cite{sun2019scalability} is a recently introduced large-scale autonomous driving dataset that would ideally be used for supervision of 3D scene flow methods. However, it does not provide dense flow annotations. While it does include all the annotations that our weakly supervised approach relies on, we are more interested in using it to evaluate the generalization capability of our method. To this end, we use the first three sequences\footnote{This results in more than 14k point cloud pairs, which is almost the same size as the whole \textit{semanticKITTI} dataset} of the \emph{waymo open} validation set and quantitatively evaluate our weakly supervised model in terms of ego-motion estimation and background segmentation. We provide qualitative results of the rigid scene flow estimation and object-level scene abstraction in \cref{fig:qualitative_results}.

Remarkably, our model that was trained only on \textit{semanticKITTI} can seamlessly generalize to \emph{waymo open}. When evaluated on the task of ego-motion estimation, it achieves an \emph{RRE} of $0.141^\circ$ and \emph{RTE} of $0.099$~m. In the BG-segmentation task, its performance on the foreground points drops slightly to $0.960$ precision and $0.689$ recall, while on the background points it remains high with $0.957$ and $0.996$ precision and recall, respectively. The drop in foreground performance can be accredited to the domain gap between the datasets~\cite{yi2020complete}; \emph{waymo open} includes many more foreground objects, especially pedestrians, than \emph{semanticKITTI}.

\vspace{-0.1cm}
\subsection{Ablation Studies}
\label{sec:ablation}
\vspace{-0.2cm}
\parahead{Influence of different loss terms}
\begin{table}[!t]
    \setlength{\tabcolsep}{6pt}
    \renewcommand{\arraystretch}{1.15}
	\centering
	\resizebox{\columnwidth}{!}{
    \begin{tabular}{cccccccc}
			\hline
			\multicolumn{3}{c}{} & \multicolumn{5}{c}{\textit{lidarKITTI}} \\
            \cline{4-8}
            $\mathcal{L}_{\text{ego}}$ & $\mathcal{L}_{\text{CD}}$ & $\mathcal{L}_{\text{rigid}}$ & EPE [m]~$\downarrow$ & Acc3DS~$\uparrow$ & Acc3DR~$\uparrow$ & RRE [$^\circ$]~$\downarrow$ & RTE [m]~$\downarrow$ \\
            \hline
            \multicolumn{1}{c}{} & \multicolumn{1}{c}{\ding{51}} & \multicolumn{1}{c}{\ding{51}} & 0.721 & 0.044 & 0.093 & 0.476 & 0.750 \\
            \multicolumn{1}{c}{\ding{51}} & \multicolumn{1}{c}{} & \multicolumn{1}{c}{\ding{51}} & 0.363  & 0.044 & 0.163 & 0.610 & 0.342 \\
            \multicolumn{1}{c}{\ding{51}} & \multicolumn{1}{c}{\ding{51}} & \multicolumn{1}{c}{} &  0.136 & 0.409 & 0.712 & 0.380 & 0.146\\
            \multicolumn{1}{c}{\ding{51}} & \multicolumn{1}{c}{\ding{51}} & \multicolumn{1}{c}{\ding{51}} & \textbf{0.134} & \textbf{0.460} & \textbf{0.746} & \textbf{0.320} & \textbf{0.130}  \\
			\hline
	\end{tabular}
	}
	\vspace{-0.1cm}
	\caption{Ablation study of the proposed training objective. All models are trained on \emph{semanticKITTI} and evaluated without test-time optimization on \emph{lidarKITTI (with ground)} dataset.}
	\vspace{-0.5cm}
	\label{tab:ablation_loss}
\end{table}
We compare our model trained with the full loss function to multiple ablations in \cref{tab:ablation_loss}. Each model is trained on \emph{semanticKITTI} and evaluated without test-time optimization on \emph{lidarKITTI} with ground points. \cref{tab:ablation_loss} shows that the terms $\mathcal{L}_\mathrm{ego}$ and $\mathcal{L}_\mathrm{CD}$ are crucial for the performance of our model. 
Adding $\mathcal{L}_\mathrm{rigid}$ further regularizes the performance and leads to an improvement in all evaluation metrics. Note how the terms that are applied only on foreground points (e.g. $\mathcal{L}_\mathrm{rigid}$) also improve the ego-motion estimation \cref{tab:ablation_loss}.
\vspace{-0.3cm}
\paragraph{Task-specific networks.}

Instead of training a single network capable of solving multiple tasks, one could also devise a combination of task-specific networks. We ablate this design choice in \cref{tab:ablation_task_specific_networks} in which we compare our full model to BG segmentation and ego-motion specific networks. Both task specific networks comprise of our backbone with the corresponding head and are trained with the $\mathcal{L}_\mathrm{BG}$ and $\mathcal{L}_\mathrm{ego}$ objective function, respectively. \cref{tab:ablation_task_specific_networks} shows that the performance of our full model is slightly inferior to the task-specific one when compared on BG segmentation. This is expected, since in our full pipeline the BG segmentation head is also trained nearly in isolation, with a single loss function. On the other hand, our full model outperforms the task-specific ego-motion model, even when the task-specific model is combined with the GT background mask. This shows that individual tasks (\eg flow and ego-motion estimation) can indeed reinforce each other, which leads to better downstream performance.
\vspace{-0.3cm}
\paragraph{Pretraining the backbone with full supervision.} We analyze the effect of initializing our weakly supervised model with pretrained backbone weights. To this end, we use the backbone weights from the model trained with full supervision in \cref{sec:experiments_backbone}. Initializing the backbone with the pretrained model leads to $1.4$~cm and $2.3$~cm improvement in terms of \emph{EPE3D} on $\emph{lidarKITTI}$ without and with ground points, respectively. Further details and additional metrics of this ablation study are available in the supplement. Note that in all evaluations presented in \cref{sec:experiments} we use the inferior model with randomly initialized weights trained only with weak supervision. 

\vspace{-4mm}
\paragraph{Run time.} 
\begin{table}[!t]
    \setlength{\tabcolsep}{4pt}
    \renewcommand{\arraystretch}{1.1}
	\centering
	\resizebox{\columnwidth}{!}{
    \begin{tabular}{cc|cccc|cc}
            \hline
            \multicolumn{2}{c|}{Task} & \multicolumn{4}{c|}{BG segmentation} & \multicolumn{2}{c}{ego-motion}\\
            
             BG seg. & ego-motion & prec. FG~$\uparrow$& rec. FG~$\uparrow$ & prec. BG~$\uparrow$ & recall. BG~$\uparrow$ & RRE [$^\circ$]~$\downarrow$ & RTE [m]~$\downarrow$ \\
            \hline
            \multicolumn{8}{c}{\textit{semanticKITTI} (w/o ground)}\\
            \hline
            \multicolumn{1}{c}{\ding{51}}  &  & \textbf{0.977} & \textbf{0.901} & \textbf{0.992} & \textbf{0.998} & - & -  \\
             &   \multicolumn{1}{c|}{\ding{51}}& - & - & - & - & 0.245 & 0.054 \\
            \multicolumn{1}{c}{\ding{51}}  &  \multicolumn{1}{c|}{\ding{51}} & 0.971 & 0.895 & 0.991 & \textbf{0.998} & \textbf{0.201} & \textbf{0.047} \\
            \hline
             \multicolumn{8}{c}{\textit{semanticKITTI} (with ground)}\\
            \hline
            \multicolumn{1}{c}{\ding{51}}  &  & \textbf{0.970} & \textbf{0.911} & \textbf{0.996} & \textbf{0.999} & - & -\\
              & \multicolumn{1}{c|}{\ding{51}}  & - & - & - & - & 0.307 & 0.071 \\
            \multicolumn{1}{c}{\ding{51}}  &\multicolumn{1}{c|}{\ding{51}}  & 0.966 & 0.904 & \textbf{0.996} & \textbf{0.999} & \textbf{0.249} & \textbf{0.059} \\
			\hline
	\end{tabular}
	}
	\vspace{-0.1cm}
	\caption{Comparison of our full pipeline with specialized networks for BG segmentation and ego-motion estimation, respectively. Note, we provide GT background masks to the ego-motion specialized network also in the test phase.}
	\label{tab:ablation_task_specific_networks}
	\vspace{-0.4cm}
\end{table}

We now compare our method to FLOT~\cite{puy20flot} and PointPWC-Net~\cite{wu2019pointpwc} in terms of run-time and number of parameters\footnote{The evaluation is performed with 8192 randomly sampled points on the \emph{lidarKITTI} dataset.}. We perform the evaluation on a standalone computer with Intel Xeon E5-1650, 32GB RAM, and a single NVIDIA Titan V. For FLOT~\cite{puy20flot} and PointPWC-Net~\cite{wu2019pointpwc} we use the official implementation provided by the authors. FLOT has the lowest number of trainable parameters ($0.11$ million) but, with $0.395$ seconds on average, also the highest run time per point cloud pair. PointPWC-Net has a larger model with approximately $7.7$ million parameters but performs one inference step in $0.147$ seconds on average. Finally, our method contains about 8 million parameters and requires $0.154$ seconds on average for a single point cloud pair. With added test-time optimization our run time increases to $0.234$ seconds on average. 
\vspace{-2mm}

\section{Conclusion}
\vspace{-0.5em}

Scene flow is the lowest level in a hierarchy of dynamic scene perception.  As such, while providing a useful cue to higher-level tasks, it is also the most demanding to supervise. Based on this observation, in this work, we have introduced a novel method that relaxes the dense supervision by integrating flow into a higher-level scene abstraction in the form of multi rigid-body motion. The result is a state-of-the-art flow estimation network that additionally outputs a concise dynamic scene representation. In particular, our mild supervision requirements are well suited for utilizing the annotation level of recently released massive data collections for autonomous driving. In future work, we plan to incorporate cues from multiple frames further seeking temporal consistency as well as increased accuracy.

\vspace{0.3em}
\small
\noindent\textbf{Acknowledgements.}
{\footnotesize This work is sponsored by Stanford-Ford Alliance, the Samsung GRO program, NSF grant IIS-1763268, the Vannevar Bush Faculty fellowship, and the NVIDIA GPU grant. We thank Barbara Verbost for her help with the visualizations and Davis Rempe for proofreading the article.}

{\small
\typeout{}
\bibliographystyle{ieee_fullname}
\bibliography{main.bib}

\begin{thebibliography}{10}\itemsep=-1pt

\bibitem{ahmed2018deep}
Eman Ahmed, Alexandre Saint, Abd El~Rahman Shabayek, Kseniya Cherenkova, Rig
  Das, Gleb Gusev, Djamila Aouada, and Bj{\"o}rn Ottersten.
\newblock Deep learning advances on different 3d data representations: A
  survey.
\newblock {\em arXiv preprint arXiv:1808.01462}, 1, 2018.

\bibitem{alhaija2015graphflow}
Hassan~Abu Alhaija, Anita Sellent, Daniel Kondermann, and Carsten Rother.
\newblock Graphflow--6d large displacement scene flow via graph matching.
\newblock In {\em German Conference on Pattern Recognition}, pages 285--296.
  Springer, 2015.

\bibitem{beauchemin1995computation}
Steven~S. Beauchemin and John~L. Barron.
\newblock The computation of optical flow.
\newblock {\em ACM computing surveys (CSUR)}, 27(3):433--466, 1995.

\bibitem{behl2019pointflownet}
Aseem Behl, Despoina Paschalidou, Simon Donn{\'e}, and Andreas Geiger.
\newblock Pointflownet: Learning representations for rigid motion estimation
  from point clouds.
\newblock In {\em Proceedings of the IEEE Conference on Computer Vision and
  Pattern Recognition}, pages 7962--7971, 2019.

\bibitem{behley2019iccv}
J. Behley, M. Garbade, A. Milioto, J. Quenzel, S. Behnke, C. Stachniss, and J.
  Gall.
\newblock {SemanticKITTI: A Dataset for Semantic Scene Understanding of LiDAR
  Sequences}.
\newblock In {\em IEEE International Conf.~on Computer Vision}, 2019.

\bibitem{besl1992icp}
Paul~J Besl and Neil~D McKay.
\newblock Method for registration of 3-d shapes.
\newblock In {\em Sensor fusion IV: control paradigms and data structures},
  volume 1611, pages 586--606. International Society for Optics and Photonics,
  1992.

\bibitem{birdal2020synchronizing}
Tolga Birdal, Michael Arbel, Umut Simsekli, and Leonidas~J Guibas.
\newblock Synchronizing probability measures on rotations via optimal
  transport.
\newblock In {\em Proceedings of the IEEE/CVF Conference on Computer Vision and
  Pattern Recognition}, pages 1569--1579, 2020.

\bibitem{birdal2018bayesian}
Tolga Birdal, Umut Simsekli, Mustafa~Onur Eken, and Slobodan Ilic.
\newblock Bayesian pose graph optimization via bingham distributions and
  tempered geodesic mcmc.
\newblock In S. Bengio, H. Wallach, H. Larochelle, K. Grauman, N. Cesa-Bianchi,
  and R. Garnett, editors, {\em Advances in Neural Information Processing
  Systems}, volume~31, pages 308--319. Curran Associates, Inc., 2018.

\bibitem{bui20206d}
Mai Bui, Tolga Birdal, Haowen Deng, Shadi Albarqouni, Leonidas Guibas, Slobodan
  Ilic, and Nassir Navab.
\newblock 6d camera relocalization in ambiguous scenes via continuous
  multimodal inference.
\newblock In Andrea Vedaldi, Horst Bischof, Thomas Brox, and Jan-Michael Frahm,
  editors, {\em Computer Vision -- ECCV 2020}, pages 139--157, Cham, 2020.
  Springer International Publishing.

\bibitem{byravan2017se3}
Arunkumar Byravan and Dieter Fox.
\newblock Se3-nets: Learning rigid body motion using deep neural networks.
\newblock In {\em 2017 IEEE International Conference on Robotics and Automation
  (ICRA)}, pages 173--180. IEEE, 2017.

\bibitem{carceroni2002multi}
Rodrigo~L Carceroni and Kiriakos~N Kutulakos.
\newblock Multi-view scene capture by surfel sampling: From video streams to
  non-rigid 3d motion, shape and reflectance.
\newblock {\em International Journal of Computer Vision}, 49(2-3):175--214,
  2002.

\bibitem{choy20194d}
Christopher Choy, JunYoung Gwak, and Silvio Savarese.
\newblock 4d spatio-temporal convnets: Minkowski convolutional neural networks.
\newblock In {\em Proceedings of the IEEE Conference on Computer Vision and
  Pattern Recognition}, 2019.

\bibitem{costeira1998multibody}
Jo{\~a}o~Paulo Costeira and Takeo Kanade.
\newblock A multibody factorization method for independently moving objects.
\newblock {\em International Journal of Computer Vision}, 29(3):159--179, 1998.

\bibitem{cuturi2013sinkhorn}
Marco Cuturi.
\newblock Sinkhorn distances: Lightspeed computation of optimal transport.
\newblock In {\em Advances in neural information processing systems}, pages
  2292--2300, 2013.

\bibitem{deng2020deep}
Haowen Deng, Mai Bui, Nassir Navab, Leonidas Guibas, Slobodan Ilic, and Tolga
  Birdal.
\newblock Deep bingham networks: Dealing with uncertainty and ambiguity in pose
  estimation.
\newblock {\em arXiv preprint arXiv:2012.11002}, 2020.

\bibitem{devernay2006multi}
Frederic Devernay, Diana Mateus, and Matthieu Guilbert.
\newblock Multi-camera scene flow by tracking 3-d points and surfels.
\newblock In {\em 2006 IEEE Computer Society Conference on Computer Vision and
  Pattern Recognition (CVPR'06)}, volume~2, pages 2203--2212. IEEE, 2006.

\bibitem{dewan2016rigid}
Ayush Dewan, Tim Caselitz, Gian~Diego Tipaldi, and Wolfram Burgard.
\newblock Rigid scene flow for 3d lidar scans.
\newblock In {\em 2016 IEEE/RSJ International Conference on Intelligent Robots
  and Systems (IROS)}, pages 1765--1770. IEEE, 2016.

\bibitem{dosovitskiy2015flownet}
Alexey Dosovitskiy, Philipp Fischer, Eddy Ilg, Philip Hausser, Caner Hazirbas,
  Vladimir Golkov, Patrick Van Der~Smagt, Daniel Cremers, and Thomas Brox.
\newblock Flownet: Learning optical flow with convolutional networks.
\newblock In {\em Proceedings of the IEEE international conference on computer
  vision}, pages 2758--2766, 2015.

\bibitem{dou2016fusion4d}
Mingsong Dou, Sameh Khamis, Yury Degtyarev, Philip Davidson, Sean~Ryan Fanello,
  Adarsh Kowdle, Sergio~Orts Escolano, Christoph Rhemann, David Kim, Jonathan
  Taylor, et~al.
\newblock Fusion4d: Real-time performance capture of challenging scenes.
\newblock {\em ACM Transactions on Graphics (TOG)}, 35(4):1--13, 2016.

\bibitem{ester1996density}
Martin Ester, Hans-Peter Kriegel, J{\"o}rg Sander, and Xiaowei Xu.
\newblock A density-based algorithm for discovering clusters in large spatial
  databases with noise.
\newblock In {\em KDD}, 1996.

\bibitem{fan2019pointrnn}
Hehe Fan and Yi Yang.
\newblock Pointrnn: Point recurrent neural network for moving point cloud
  processing.
\newblock {\em arXiv preprint arXiv:1910.08287}, 2019.

\bibitem{Geiger2012CVPR}
Andreas Geiger, Philip Lenz, and Raquel Urtasun.
\newblock Are we ready for autonomous driving? the kitti vision benchmark
  suite.
\newblock In {\em Conference on Computer Vision and Pattern Recognition
  (CVPR)}, 2012.

\bibitem{golyanik2017multiframe}
Vladislav Golyanik, Kihwan Kim, Robert Maier, Matthias Nie{\ss}ner, Didier
  Stricker, and Jan Kautz.
\newblock Multiframe scene flow with piecewise rigid motion.
\newblock In {\em 2017 International Conference on 3D Vision (3DV)}, pages
  273--281. IEEE, 2017.

\bibitem{gu2019hplflownet}
Xiuye Gu, Yijie Wang, Chongruo Wu, Yong~Jae Lee, and Panqu Wang.
\newblock Hplflownet: Hierarchical permutohedral lattice flownet for scene flow
  estimation on large-scale point clouds.
\newblock In {\em IEEE Conference on Computer Vision and Pattern Recognition},
  pages 3254--3263, 2019.

\bibitem{he2017mask}
Kaiming He, Georgia Gkioxari, Piotr Doll{\'a}r, and Ross Girshick.
\newblock Mask r-cnn.
\newblock In {\em Proceedings of the IEEE international conference on computer
  vision}, pages 2961--2969, 2017.

\bibitem{horn1981determining}
Berthold~KP Horn and Brian~G Schunck.
\newblock Determining optical flow.
\newblock In {\em Techniques and Applications of Image Understanding}, volume
  281, pages 319--331. International Society for Optics and Photonics, 1981.

\bibitem{hornacek2014sphereflow}
Michael Hornacek, Andrew Fitzgibbon, and Carsten Rother.
\newblock Sphereflow: 6 dof scene flow from rgb-d pairs.
\newblock In {\em Proceedings of the IEEE Conference on Computer Vision and
  Pattern Recognition}, pages 3526--3533, 2014.

\bibitem{huang2019clusterslam}
Jiahui Huang, Sheng Yang, Zishuo Zhao, Yu-Kun Lai, and Shi-Min Hu.
\newblock Clusterslam: A slam backend for simultaneous rigid body clustering
  and motion estimation.
\newblock In {\em Proceedings of the IEEE International Conference on Computer
  Vision}, pages 5875--5884, 2019.

\bibitem{huguet2007variational}
Fr{\'e}d{\'e}ric Huguet and Fr{\'e}d{\'e}ric Devernay.
\newblock A variational method for scene flow estimation from stereo sequences.
\newblock In {\em 2007 IEEE 11th International Conference on Computer Vision},
  pages 1--7. IEEE, 2007.

\bibitem{ilg2017flownet}
Eddy Ilg, Nikolaus Mayer, Tonmoy Saikia, Margret Keuper, Alexey Dosovitskiy,
  and Thomas Brox.
\newblock Flownet 2.0: Evolution of optical flow estimation with deep networks.
\newblock In {\em Proceedings of the IEEE conference on computer vision and
  pattern recognition}, pages 2462--2470, 2017.

\bibitem{innmann2016volumedeform}
Matthias Innmann, Michael Zollh{\"o}fer, Matthias Nie{\ss}ner, Christian
  Theobalt, and Marc Stamminger.
\newblock Volumedeform: Real-time volumetric non-rigid reconstruction.
\newblock In {\em European Conference on Computer Vision}, pages 362--379.
  Springer, 2016.

\bibitem{jaimez2017fast}
Mariano Jaimez, Christian Kerl, Javier Gonzalez-Jimenez, and Daniel Cremers.
\newblock Fast odometry and scene flow from rgb-d cameras based on geometric
  clustering.
\newblock In {\em 2017 IEEE International Conference on Robotics and Automation
  (ICRA)}, pages 3992--3999. IEEE, 2017.

\bibitem{jaimez15icra}
M. Jaimez, M. Souiai, J. Gonzalez-Jimenez, and D. Cremers.
\newblock A primal-dual framework for real-time dense rgb-d scene flow.
\newblock In {\em Proc. of the IEEE Int. Conf. on Robotics and Automation
  (ICRA)}, 2015.

\bibitem{jiang2019sense}
Huaizu Jiang, Deqing Sun, Varun Jampani, Zhaoyang Lv, Erik Learned-Miller, and
  Jan Kautz.
\newblock Sense: A shared encoder network for scene-flow estimation.
\newblock In {\em Proceedings of the IEEE International Conference on Computer
  Vision}, pages 3195--3204, 2019.

\bibitem{jiang2020pointgroup}
Li Jiang, Hengshuang Zhao, Shaoshuai Shi, Shu Liu, Chi-Wing Fu, and Jiaya Jia.
\newblock Pointgroup: Dual-set point grouping for 3d instance segmentation.
\newblock In {\em Proceedings of the IEEE/CVF Conference on Computer Vision and
  Pattern Recognition}, pages 4867--4876, 2020.

\bibitem{kabsch1976solution}
Wolfgang Kabsch.
\newblock A solution for the best rotation to relate two sets of vectors.
\newblock {\em Acta Crystallographica Section A: Crystal Physics, Diffraction,
  Theoretical and General Crystallography}, 32(5):922--923, 1976.

\bibitem{kanatani2001motion}
Ken-ichi Kanatani.
\newblock Motion segmentation by subspace separation and model selection.
\newblock In {\em Proceedings Eighth IEEE International Conference on computer
  Vision. ICCV 2001}, volume~2, pages 586--591. IEEE, 2001.

\bibitem{kingma2014adam}
Diederik~P Kingma and Jimmy Ba.
\newblock Adam: A method for stochastic optimization.
\newblock {\em arXiv preprint arXiv:1412.6980}, 2014.

\bibitem{kullback1951information}
Solomon Kullback and Richard~A Leibler.
\newblock On information and sufficiency.
\newblock {\em The annals of mathematical statistics}, 22(1):79--86, 1951.

\bibitem{liu2019flownet3d}
Xingyu Liu, Charles~R Qi, and Leonidas~J Guibas.
\newblock Flownet3d: Learning scene flow in 3d point clouds.
\newblock In {\em Proceedings of the IEEE Conference on Computer Vision and
  Pattern Recognition}, pages 529--537, 2019.

\bibitem{liu2019meteornet}
Xingyu Liu, Mengyuan Yan, and Jeannette Bohg.
\newblock Meteornet: Deep learning on dynamic 3d point cloud sequences.
\newblock In {\em IEEE International Conference on Computer Vision}, pages
  9246--9255, 2019.

\bibitem{lv2018learning}
Zhaoyang Lv, Kihwan Kim, Alejandro Troccoli, Deqing Sun, James~M Rehg, and Jan
  Kautz.
\newblock Learning rigidity in dynamic scenes with a moving camera for 3d
  motion field estimation.
\newblock In {\em Proceedings of the European Conference on Computer Vision
  (ECCV)}, pages 468--484, 2018.

\bibitem{ma2019deep}
Wei-Chiu Ma, Shenlong Wang, Rui Hu, Yuwen Xiong, and Raquel Urtasun.
\newblock Deep rigid instance scene flow.
\newblock In {\em Proceedings of the IEEE Conference on Computer Vision and
  Pattern Recognition}, pages 3614--3622, 2019.

\bibitem{mayer2016FT3D}
N. Mayer, E. Ilg, P. H{\"a}usser, P. Fischer, D. Cremers, A. Dosovitskiy, and
  T. Brox.
\newblock A large dataset to train convolutional networks for disparity,
  optical flow, and scene flow estimation.
\newblock In {\em IEEE International Conference on Computer Vision and Pattern
  Recognition (CVPR)}, 2016.
\newblock arXiv:1512.02134.

\bibitem{menze2015joint}
Moritz Menze, Christian Heipke, and Andreas Geiger.
\newblock Joint 3d estimation of vehicles and scene flow.
\newblock {\em ISPRS Annals of Photogrammetry, Remote Sensing \& Spatial
  Information Sciences}, 2, 2015.

\bibitem{menze2018object}
Moritz Menze, Christian Heipke, and Andreas Geiger.
\newblock Object scene flow.
\newblock {\em ISPRS Journal of Photogrammetry and Remote Sensing}, 140:60--76,
  2018.

\bibitem{mittal2019just}
Himangi Mittal, Brian Okorn, and David Held.
\newblock Just go with the flow: Self-supervised scene flow estimation.
\newblock {\em arXiv preprint arXiv:1912.00497}, 2019.

\bibitem{moosmann2013joint}
Frank Moosmann and Christoph Stiller.
\newblock Joint self-localization and tracking of generic objects in 3d range
  data.
\newblock In {\em 2013 IEEE International Conference on Robotics and
  Automation}, pages 1146--1152. IEEE, 2013.

\bibitem{mustafa2019semantically}
Armin Mustafa and Adrian Hilton.
\newblock Semantically coherent 4d scene flow of dynamic scenes.
\newblock {\em International Journal of Computer Vision}, pages 1--17, 2019.

\bibitem{nair2010rectified}
Vinod Nair and Geoffrey~E Hinton.
\newblock Rectified linear units improve restricted boltzmann machines.
\newblock In {\em ICML}, 2010.

\bibitem{newcombe2015dynamicfusion}
Richard~A Newcombe, Dieter Fox, and Steven~M Seitz.
\newblock Dynamicfusion: Reconstruction and tracking of non-rigid scenes in
  real-time.
\newblock In {\em Proceedings of the IEEE conference on computer vision and
  pattern recognition}, pages 343--352, 2015.

\bibitem{niemeyer2019occupancy}
Michael Niemeyer, Lars Mescheder, Michael Oechsle, and Andreas Geiger.
\newblock Occupancy flow: 4d reconstruction by learning particle dynamics.
\newblock In {\em Proceedings of the IEEE International Conference on Computer
  Vision}, pages 5379--5389, 2019.

\bibitem{pedregosa2011scikit}
F. Pedregosa, G. Varoquaux, A. Gramfort, V. Michel, B. Thirion, O. Grisel, M.
  Blondel, P. Prettenhofer, R. Weiss, V. Dubourg, J. Vanderplas, A. Passos, D.
  Cournapeau, M. Brucher, M. Perrot, and E. Duchesnay.
\newblock Scikit-learn: Machine learning in {P}ython.
\newblock {\em Journal of Machine Learning Research}, 12:2825--2830, 2011.

\bibitem{pons2005modelling}
J-P Pons, Renaud Keriven, and Olivier Faugeras.
\newblock Modelling dynamic scenes by registering multi-view image sequences.
\newblock In {\em 2005 IEEE Computer Society Conference on Computer Vision and
  Pattern Recognition (CVPR'05)}, volume~2, pages 822--827. IEEE, 2005.

\bibitem{puy20flot}
Gilles Puy, Alexandre Boulch, and Renaud Marlet.
\newblock {FLOT}: {S}cene {F}low on {P}oint {C}louds {G}uided by {O}ptimal
  {T}ransport.
\newblock In {\em European Conference on Computer Vision}, 2020.

\bibitem{qi2017pointnet}
Charles~R Qi, Hao Su, Kaichun Mo, and Leonidas~J Guibas.
\newblock Pointnet: Deep learning on point sets for 3d classification and
  segmentation.
\newblock In {\em IEEE conference on computer vision and pattern recognition},
  pages 652--660, 2017.

\bibitem{qi2017pointnet++}
Charles~Ruizhongtai Qi, Li Yi, Hao Su, and Leonidas~J Guibas.
\newblock Pointnet++: Deep hierarchical feature learning on point sets in a
  metric space.
\newblock In {\em Advances in neural information processing systems}, pages
  5099--5108, 2017.

\bibitem{rempe2020caspr}
Davis Rempe, Tolga Birdal, Yongheng Zhao, Zan Gojcic, Srinath Sridhar, and
  Leonidas~J. Guibas.
\newblock Caspr: Learning canonical spatiotemporal point cloud representations.
\newblock In {\em Advances in Neural Information Processing Systems (NeurIPS)},
  2020.

\bibitem{ronneberger2015u}
Olaf Ronneberger, Philipp Fischer, and Thomas Brox.
\newblock U-net: Convolutional networks for biomedical image segmentation.
\newblock In {\em International Conference on Medical image computing and
  computer-assisted intervention}, pages 234--241. Springer, 2015.

\bibitem{runz2018maskfusion}
Martin Runz, Maud Buffier, and Lourdes Agapito.
\newblock Maskfusion: Real-time recognition, tracking and reconstruction of
  multiple moving objects.
\newblock In {\em 2018 IEEE International Symposium on Mixed and Augmented
  Reality (ISMAR)}, pages 10--20. IEEE, 2018.

\bibitem{salti2011performance}
Samuele Salti, Federico Tombari, and Luigi Di~Stefano.
\newblock A performance evaluation of 3d keypoint detectors.
\newblock In {\em 2011 International Conference on 3D Imaging, Modeling,
  Processing, Visualization and Transmission}, pages 236--243. IEEE, 2011.

\bibitem{saputra2018visual}
Muhamad Risqi~U Saputra, Andrew Markham, and Niki Trigoni.
\newblock Visual slam and structure from motion in dynamic environments: A
  survey.
\newblock {\em ACM Computing Surveys (CSUR)}, 51(2):1--36, 2018.

\bibitem{sinkhorn1964relationship}
Richard Sinkhorn.
\newblock A relationship between arbitrary positive matrices and doubly
  stochastic matrices.
\newblock {\em The annals of mathematical statistics}, 35(2):876--879, 1964.

\bibitem{sinkhorn1967diagonal}
Richard Sinkhorn.
\newblock Diagonal equivalence to matrices with prescribed row and column sums.
\newblock {\em The American Mathematical Monthly}, 74(4):402--405, 1967.

\bibitem{slavcheva2017killingfusion}
Miroslava Slavcheva, Maximilian Baust, Daniel Cremers, and Slobodan Ilic.
\newblock Killingfusion: Non-rigid 3d reconstruction without correspondences.
\newblock In {\em Proceedings of the IEEE Conference on Computer Vision and
  Pattern Recognition}, pages 1386--1395, 2017.

\bibitem{slavcheva2018sobolevfusion}
Miroslava Slavcheva, Maximilian Baust, and Slobodan Ilic.
\newblock Sobolevfusion: 3d reconstruction of scenes undergoing free non-rigid
  motion.
\newblock In {\em Proceedings of the IEEE Conference on Computer Vision and
  Pattern Recognition}, pages 2646--2655, 2018.

\bibitem{strecke2019fusion}
Michael Strecke and Jorg Stuckler.
\newblock Em-fusion: Dynamic object-level slam with probabilistic data
  association.
\newblock In {\em Proceedings of the IEEE International Conference on Computer
  Vision}, pages 5865--5874, 2019.

\bibitem{su2018splatnet}
Hang Su, Varun Jampani, Deqing Sun, Subhransu Maji, Evangelos Kalogerakis,
  Ming-Hsuan Yang, and Jan Kautz.
\newblock Splatnet: Sparse lattice networks for point cloud processing.
\newblock In {\em Proceedings of the IEEE Conference on Computer Vision and
  Pattern Recognition}, pages 2530--2539, 2018.

\bibitem{sun2019scalability}
Pei Sun, Henrik Kretzschmar, Xerxes Dotiwalla, Aurelien Chouard, Vijaysai
  Patnaik, Paul Tsui, James Guo, Yin Zhou, Yuning Chai, Benjamin Caine, et~al.
\newblock Scalability in perception for autonomous driving: An open dataset
  benchmark.
\newblock {\em arXiv preprint arXiv:1912.04838}, 2019.

\bibitem{tanzmeister2014grid}
Georg Tanzmeister, Julian Thomas, Dirk Wollherr, and Martin Buss.
\newblock Grid-based mapping and tracking in dynamic environments using a
  uniform evidential environment representation.
\newblock In {\em 2014 IEEE International Conference on Robotics and Automation
  (ICRA)}, pages 6090--6095. IEEE, 2014.

\bibitem{tchapmi2017segcloud}
Lyne Tchapmi, Christopher Choy, Iro Armeni, JunYoung Gwak, and Silvio Savarese.
\newblock Segcloud: Semantic segmentation of 3d point clouds.
\newblock In {\em 2017 international conference on 3D vision (3DV)}, pages
  537--547. IEEE, 2017.

\bibitem{tishchenko2020selfflow}
Ivan Tishchenko, Sandro Lombardi, Martin~R Oswald, and Marc Pollefeys.
\newblock Self-supervised learning of non-rigid residual flow and ego-motion.
\newblock {\em arXiv preprint arXiv:2009.10467}, 2020.

\bibitem{ushani2017learning}
Arash~K Ushani, Ryan~W Wolcott, Jeffrey~M Walls, and Ryan~M Eustice.
\newblock A learning approach for real-time temporal scene flow estimation from
  lidar data.
\newblock In {\em 2017 IEEE International Conference on Robotics and Automation
  (ICRA)}, pages 5666--5673. IEEE, 2017.

\bibitem{vedula1999three}
Sundar Vedula, Simon Baker, Peter Rander, Robert Collins, and Takeo Kanade.
\newblock Three-dimensional scene flow.
\newblock In {\em IEEE International Conference on Computer Vision}, pages
  722--729. IEEE, 1999.

\bibitem{vogel2013piecewise}
Christoph Vogel, Konrad Schindler, and Stefan Roth.
\newblock Piecewise rigid scene flow.
\newblock In {\em Proceedings of the IEEE International Conference on Computer
  Vision}, pages 1377--1384, 2013.

\bibitem{wang2018deep}
Shenlong Wang, Simon Suo, Wei-Chiu Ma, Andrei Pokrovsky, and Raquel Urtasun.
\newblock Deep parametric continuous convolutional neural networks.
\newblock In {\em IEEE Conference on Computer Vision and Pattern Recognition},
  pages 2589--2597, 2018.

\bibitem{wang2018sgpn}
Weiyue Wang, Ronald Yu, Qiangui Huang, and Ulrich Neumann.
\newblock Sgpn: Similarity group proposal network for 3d point cloud instance
  segmentation.
\newblock In {\em IEEE Conference on Computer Vision and Pattern Recognition},
  pages 2569--2578, 2018.

\bibitem{wang2019flownet3d++}
Zirui Wang, Shuda Li, Henry Howard-Jenkins, Victor~Adrian Prisacariu, and Min
  Chen.
\newblock Flownet3d++: Geometric losses for deep scene flow estimation.
\newblock {\em arXiv preprint arXiv:1912.01438}, 2019.

\bibitem{Wedel-et-al-08}
A. Wedel, C. Rabe, T. Vaudrey, T. Brox, U. Franke, and D. Cremers.
\newblock Efficient dense scene flow from sparse or dense stereo data.
\newblock In {\em ECCV}, Marseille, France, October 2008.

\bibitem{wu2019pointpwc}
Wenxuan Wu, Zhiyuan Wang, Zhuwen Li, Wei Liu, and Li Fuxin.
\newblock Pointpwc-net: A coarse-to-fine network for supervised and
  self-supervised scene flow estimation on 3d point clouds.
\newblock {\em arXiv preprint arXiv:1911.12408}, 2019.

\bibitem{xie2020pointcontrast}
Saining Xie, Jiatao Gu, Demi Guo, Charles~R Qi, Leonidas~J Guibas, and Or
  Litany.
\newblock Pointcontrast: Unsupervised pre-training for 3d point cloud
  understanding.
\newblock {\em arXiv preprint arXiv:2007.10985}, 2020.

\bibitem{yang2018toward}
Jiaqi Yang, Yang Xiao, and Zhiguo Cao.
\newblock Toward the repeatability and robustness of the local reference frame
  for 3d shape matching: An evaluation.
\newblock {\em IEEE Transactions on Image Processing}, 27(8):3766--3781, 2018.

\bibitem{yang2019modeling}
Jiancheng Yang, Qiang Zhang, Bingbing Ni, Linguo Li, Jinxian Liu, Mengdie Zhou,
  and Qi Tian.
\newblock Modeling point clouds with self-attention and gumbel subset sampling.
\newblock In {\em IEEE Conference on Computer Vision and Pattern Recognition},
  pages 3323--3332, 2019.

\bibitem{yew2020rpm}
Zi~Jian Yew and Gim~Hee Lee.
\newblock Rpm-net: Robust point matching using learned features.
\newblock In {\em CVPR}, 2020.

\bibitem{yi2020complete}
Li Yi, Boqing Gong, and Thomas Funkhouser.
\newblock Complete \& label: A domain adaptation approach to semantic
  segmentation of lidar point clouds, 2020.

\bibitem{yi2019deep}
Li Yi, Haibin Huang, Difan Liu, Evangelos Kalogerakis, Hao Su, and Leonidas
  Guibas.
\newblock Deep part induction from articulated object pairs.
\newblock {\em ACM Transactions on Graphics (TOG)}, 37(6):209, 2019.

\bibitem{yi2019gspn}
Li Yi, Wang Zhao, He Wang, Minhyuk Sung, and Leonidas~J Guibas.
\newblock Gspn: Generative shape proposal network for 3d instance segmentation
  in point cloud.
\newblock In {\em IEEE Conference on Computer Vision and Pattern Recognition},
  pages 3947--3956, 2019.

\bibitem{zaheer2018deep}
Manzil Zaheer, Satwik Kottur, Siamak Ravanbakhsh, Barnabas Poczos, Ruslan
  Salakhutdinov, and Alexander Smola.
\newblock Deep sets, 2018.

\bibitem{zamir2020robust}
Amir~R Zamir, Alexander Sax, Nikhil Cheerla, Rohan Suri, Zhangjie Cao, Jitendra
  Malik, and Leonidas~J Guibas.
\newblock Robust learning through cross-task consistency.
\newblock In {\em Proceedings of the IEEE/CVF Conference on Computer Vision and
  Pattern Recognition}, pages 11197--11206, 2020.

\bibitem{zamir2018taskonomy}
Amir~R Zamir, Alexander Sax, William Shen, Leonidas~J Guibas, Jitendra Malik,
  and Silvio Savarese.
\newblock Taskonomy: Disentangling task transfer learning.
\newblock In {\em Proceedings of the IEEE conference on computer vision and
  pattern recognition}, pages 3712--3722, 2018.

\bibitem{Zhou2018}
Qian-Yi Zhou, Jaesik Park, and Vladlen Koltun.
\newblock {Open3D}: {A} modern library for {3D} data processing.
\newblock {\em arXiv:1801.09847}, 2018.

\bibitem{zhou2018voxelnet}
Yin Zhou and Oncel Tuzel.
\newblock Voxelnet: End-to-end learning for point cloud based 3d object
  detection.
\newblock In {\em Proceedings of the IEEE Conference on Computer Vision and
  Pattern Recognition}, pages 4490--4499, 2018.

\end{thebibliography}
}
\setcounter{section}{0}
\renewcommand\thesection{\Alph{section}}
\newcommand{\suppsection}{\subsection}
\section{Appendix}

\insertimageStar{1}{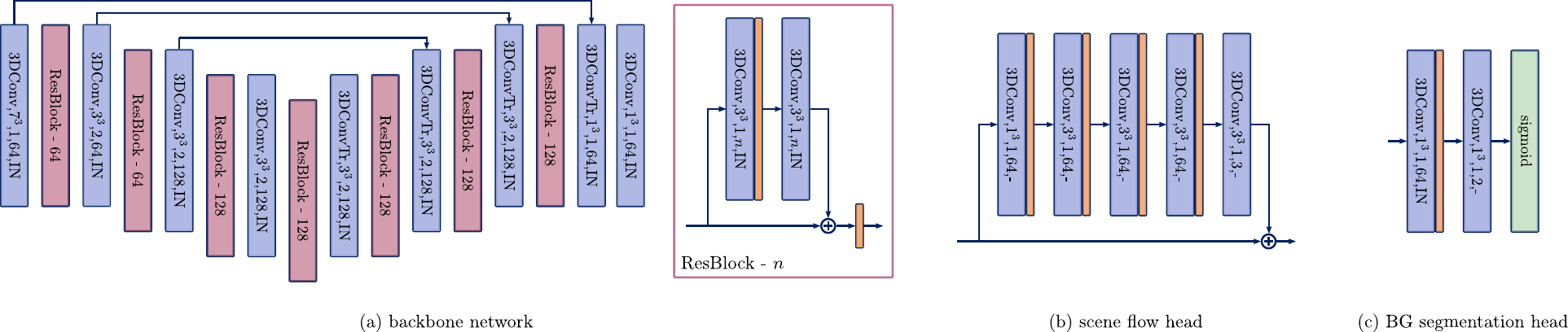}{Network architecture of the backbone network (a), scene flow head (b), and background segmentation head (c). Blue blocks denote the convolutional (3DConv) and transpose convolutational (3DConvTr) layers, where \emph{3DConv,a,b,c,d} denotes a 3D convolutional layer with the kernel size $a$, stride $b$, output feature dimension $c$, and normalization function $d$. Orange rectangle denotes the ReLU activation function~\cite{nair2010rectified} and IN is the instance normalization.\vspace{-3mm} }{fig:network_arch}{t!}

\subsection{Implementation details}
We now provide the implementation details of our network architecture, foreground clustering using DBSCAN, and test-time optimization scheme. Additionally, we summarize the Kabsch algorithm and the entropy regularized optimal transport. For a general overview please refer to~\cref{sec:method} and \cref{fig:SFpipeline}.

\subsubsection{Network architecture}
\label{sec:network_architecture}
Our whole network is built upon MinkowskiEngine~\cite{choy20194d}, an auto-differentiation library for sparse tensors. \review{In all experiments, we first randomly sample 8192 points from the source and target point cloud, respectively.} We then voxelize the original point clouds into sparse tensors with a voxel size of $0.1$~m. If more than 8192 voxels remain, we randomly select 8192 of them. For the evaluation, all the inferred quantities are transferred from the voxels to originally sampled points using \cref{eq:voxel_to_points}. The combined network has 8,078,149 learnable parameters. \cref{fig:network_arch} shows the detailed architecture, where blue rectangles denote the Minkowski convolutional and transpose convolutional layers, violet rectangles the ResBlocks, and orange rectangles are ReLU activation functions.

\parahead{Backbone network}
The detailed architecture of the backbone network is depicted in \cref{fig:network_arch} (a). We use absolute coordinates instead of the more common binary occupancy as the input feature of the first layer. All convolutional layers are followed by instance normalization and in ResBlocks additionally by the ReLU activation function. The output of the backbone network are 64-dimensional pointwise latent features. Due to the fixed voxel size, the number of points can vary between the point clouds.

\parahead{Scene flow head}
The scene flow head, depicted in \cref{fig:network_arch} (b), takes the initial flow vector field $\V^\mathrm{init}$ as input and computes a residual flow $\Delta\V^\mathrm{init}$. All layers except the last are followed by the ReLU activation function. Normalization functions are not used in the scene flow head.

\parahead{Background segmentation head}
The background segmentation head (\cref{fig:network_arch} (c)) comprises two sparse convolutional layers, where the first one is followed by the instance normalization and ReLU activation function. The output of the second layer is passed through a sigmoid function to obtain the foreground probabilities $\mathbf{h}$.

\subsubsection{\review{Foreground clustering}}
\review{We cluster the foreground points into objects using a simple DBSCAN clustering algorithm. Specifically, we first extract the foreground points $\mathbf{X}^f$ of the source point cloud using the inferred foreground probabilities $\mathbf{h}^\mathbf{X}$. We then run a scikit-learn~\cite{pedregosa2011scikit} implementation of DBSCAN with $eps = 0.75$~m and the minimum number of samples in the neighborhood equal to 5. After clustering, we only retain clusters with at least 10 points. Our clustering algorithm is based on the hypothesis that the objects scattered in the scene, are separated by void space. If two objects (e.g. cars) are too close to each other and get assigned to the same cluster, our method does not have the capacity to recover from this wrong assignment. However, we conduct an ablation study (\cref{sec:additional_results}) and empirically confirm the above-mentioned hypothesis for the autonomous driving datasets.}

\subsubsection{Test-time optimization}
\label{sec:optimization_scheme}
The object-level abstraction of the scene enables us to perform test-time optimization of per-object rigid body transformations. This optimization is performed independently for the background and each of the foreground objects. In the following we assume that our network outputs object-level masks $\{\mathbf{z}\}_{k=1}^K$ and transformation parameters $\{\T_k\}_{k=1}^K$.

\parahead{Background transformation}
Let $\mathbf{z}_1$ and $\T_1$ denote the inferred background mask and ego-motion transformation parameters, respectively. We use $\mathbf{z}_1$ to extract the background points $\X^b$ and $\Y^b$ of the source and target point cloud, transform $\X^b$ with the inferred ego-motion $\T_1$ and use these point clouds as an input to the ICP~\cite{besl1992icp} optimization of the transformation parameters. Specifically, we use the Open3D~\cite{Zhou2018} implementation with the maximum correspondence distance of $0.15$~m and a maximum number of iterations equal to $300$. On \emph{lidarKITTI} the optimization of the background transformation parameters takes $0.06$~s on average for a point cloud pair, using the same computer as in the run time experiments in \cref{sec:ablation}. 

\parahead{Object level transformations}
For the object level transformations, we follow a similar procedure. We start by extracting the points $\X^k$ of the object $k$ in the source point cloud using the object mask $\mathbf{z}_k$, and all the foreground points $\Y^f$ of the target point cloud using the complementary mask of the background mask $\mathbf{z}_1$. We then transform the points $\X^k$ with the inferred transformation parameters $\T_k$ and again use the point clouds as input to the ICP optimization of the transformation parameters. %
We use all the foreground points of the target point cloud as we do not have access to the instance level correspondences. The optimization of all foreground transformation parameters for a single \emph{lidarKITTI} point cloud pair takes approximately $0.02$~s on average. We perform at most 300 iterations of ICP per object with a maximum correspondences distance equal to $0.25$~m.
\subsubsection{Kabsch algorithm}
\label{sec:kabsch}
\review{In order to make the paper self contained, we now summarize the weighted Kabsch algorithm, which represents the closed-form differentiable solution to the ego-motion estimation problem:}
\begin{equation}
    \Rego^\star, \tego^\star = \argmin_{\Rego,\tego} \sum_{l=1}^{N^b}  w_l \| \Rego\x^b_l + \tego - \phi({\x^b_l,\Y^b}) \|^2.
    \label{eq:weighted_pairwise}\nonumber
\end{equation}
\review{In the following, we omit the superscript $b$ and define $\mathbf{q}_l := \phi({\x_l,\Y})$ and the resulting point cloud of correspondences as $\mathbf{Q} \in  \mathbb{R}^{3 \times N} = \{\mathbf{q}_l \in \mathbb{R}^3\}_l$. Let $\overline{\mathbf{x}}$ and $\overline{\mathbf{q}}$: 
\begin{equation}
    \overline{\mathbf{x}} := \frac{\sum^N_{l=1} w_l \mathbf{x}_l}{\sum^N_{l=1} w_l}, \quad
    \overline{\mathbf{q}} := \frac{\sum^N_{l=1} w_l \mathbf{q}_l}{\sum^N_{l=1} w_l}
\end{equation}
denote the weighted centroids of point clouds $\mathbf{X}$ and $\mathbf{Q}$, respectively. The centered point coordinates can then be computed as $    \widetilde{\mathbf{x}}_l := \mathbf{x}_l - \mathbf{\overline{\mathbf{x}}}$ and $\widetilde{\mathbf{q}}_l := \mathbf{q}_l - \mathbf{\overline{\mathbf{q}}}$. By  arranging them back to the matrix form $\widetilde{\mathbf{X}}  \in \mathbb{R}^{N \times 3}$ and $\widetilde{\mathbf{Q}}  \in \mathbb{R}^{N \times 3}$, a weighted covariance matrix $\mathbf{S} \in \mathbb{R}^{3 \times 3}$ can be computed as 
\begin{equation}
    \mathbf{S} = \widetilde{\mathbf{X}}^T\mathbf{W}\widetilde{\mathbf{Q}}
\end{equation}
where $\mathbf{W} = \text{diag}(w_1,\ldots,w_{N})$. Considering the singular value decomposition $\mathbf{S}=\mathbf{U}\mathbf{\Sigma}\mathbf{V}^T$, the optimal rotation matrix is given by
\begin{equation}
     \Rego^\star = \mathbf{V}\begin{bmatrix*}[c] 
    1 & 0 & 0 \\
    0 & 1 & 0 \\
    0 & 0 & \text{det}(\mathbf{V}\mathbf{U}^T)\\ \end{bmatrix*}\mathbf{U}^T  
\end{equation}
where $\text{det}(\cdot)$ denotes computing the determinant and is used to avoid generating a reflection matrix. Finally, the optimal translation vector is recovered as
\begin{equation}
    \tego^\star = \overline{\mathbf{q}} - \Rego^\star\overline{\mathbf{x}}.
\end{equation}
Note, when estimating the per-object transformation parameters of the foreground points, we use an unweighted Kabsch algorithm, \ie $\mathbf{W}=\Id$.}
\subsubsection{Entropy-regularized optimal transport}
We provide a brief overview of the Sinkhorn algorithm used to approximate the optimal transport following~\cite{birdal2020synchronizing}. 
Consider a discrete probability measure $\mur$ on the simplex $\Sigma_d\triangleq\{\x \in \R_+^n : \x^\top\one_n=1\}$ with weights $\ab=\{a_i\}$ and locations $\{x_i\}, i=1\dots n$ as:
\begin{align}
    \mur = \sum\nolimits_{i=1}^n a_i \delta_{x_i}, \quad a_i\geq 0 \,\,\wedge\,\, \sum\nolimits_{i=1}^n a_i = 1 
    \label{eqn:discrete_meas}
\end{align}
where $\delta_{x}$ is a Dirac delta function at $x$. 

For two probability measures $\mur$ and $\muc$ in the simplex, let $U(\mur,\muc)$ denote the polyhedral set of $n_{\mur}\times n_{\muc}$ matrices:
\begin{align*}
    U(\mur,\muc) = \{\Pb\in\R^{n_{\mur}\times n_{\muc}}_{+} \,:\, \Pb\one_{n_{\muc}} = \mur \,\wedge\, \Pb^\top \one_{n_{\mur}} = \muc \}
\end{align*}
where $\one_d$ is a $d$-dimensional vector of ones. $U(\mur,\muc)$ is also referred to as the \emph{transportation polytope}. 

Let $\C$ be a $n_{\mur}\times n_{\muc}$ cost matrix that is constructed from the \emph{ground cost} function $c(x_i^{\mur}, x_j^{\muc}) $. Kantorovich's optimal transport formulation seeks to find the transport plan optimally mapping $\mur$ to $\muc$:
\begin{align}\label{eq:WK}
    \mathcal{W}_c(\mur, \muc) = d_{c}(\mur, \muc) = \min\nolimits_{\Pb\in U(\mur,\muc)} \langle \Pb, \C \rangle 
\end{align}
with $\langle\cdot\rangle$ denoting the Frobenius dot-product. Note that this expression is also known as the \textit{Wasserstein distance} (WD).
Due to the computational difficulties in minimizing~\cref{eq:WK}
Cuturi~\cite{cuturi2013sinkhorn} introduced an alternative convex set consisting of joint probabilities with small enough mutual information:
\begin{align}
    U_{\alpha}(\mur,\muc) = \{\Pb\in U(\mur,\muc) \,:\, \emph{KL}\infdivx{\Pb} {\mur\muc^\top}\leq \alpha \}
\end{align}
where $\emph{KL}(\cdot)$ refers to the Kullback Leibler divergence~\cite{kullback1951information}. This restricted polytope leads to a tractable distance between discrete probability measures under the cost matrix $\C$ constructed from the function $c$:
\begin{align}
    d_{c,\alpha}(\mur, \muc) = \min\nolimits_{\Pb \in U_{\alpha}(\mur,\muc)} \langle \Pb, \C \rangle.
\end{align}
$d_{c,\alpha}$ can be computed by a modification of simple matrix scaling algorithms, such as Sinkhorn's algorithm~\cite{sinkhorn1964relationship,sinkhorn1967diagonal}. Note that here the cost matrix is inversely related to the affinity matrix defined in the main paper.

\subsection{Datasets}
\label{sec:datasets}
In this work, a total of five datasets are used to train and evaluate the performance of the proposed approach. In the following, we detail the generation of the training and evaluation data. For all datasets, we use the same coordinate system centered at the camera or LiDAR sensor. The $z$-axis of the coordinate system points in the viewing direction of the camera or to the front of the car, the positive $y$-axis points in the up direction, and $x$ completes the right-handed coordinate system. All processed point clouds are available under \url{https://3dsceneflow.github.io/}. 

\parahead{FlyingThings3D~\cite{mayer2016FT3D}} FlyingThings3D is a large-scale synthetic dataset of stereo RGB images proposed to train 2D based scene flow networks. We follow~\cite{gu2019hplflownet} and use the subset of the dataset in which the extremely hard examples are omitted. The point clouds are obtained by lifting the stereo images to 3D using the annotated disparity maps, optical flow, and camera parameters. We again follow \cite{gu2019hplflownet} and remove the points whose disparity and optical flow are occluded (occlusion maps are part of the original dataset). Additionally, points with a depth larger than 35~m are removed~\cite{gu2019hplflownet}. The resulting point clouds are in direct correspondence, i.e. $\Y = \X + \V$, where $\X$ and $\Y$ are the source and target point cloud, and $\V$ is the ground truth flow. For training and evaluation, 8192 points are randomly sampled from each point cloud independently.

\parahead{stereoKITTI~\cite{menze2015joint,menze2018object}} This is a dataset based on the KITTI Scene Flow benchmark, which is designed for the evaluation of stereo based scene flow methods.  It consists of 200 training and 200 test scenes. Because the ground truth annotations of the test scenes are not available, only the training scenes are used for evaluation. Due to incomplete and noisy annotations, 58 scenes are further removed and the final evaluation set comprises 142 scenes~\cite{liu2019flownet3d, gu2019hplflownet}. The stereo images are lifted to point clouds analogously to FlyingThings3D~\cite{gu2019hplflownet}. Again, occluded points and points with a depth larger than $35$~m are removed. For evaluation, we sample 8192 points from each point cloud independently. \cref{tab:flow_traditional} shows the result of the typical evaluation protocol in which the ground points are removed by a naive thresholding the $y$ coordinate at $-1.4$~m~\cite{liu2019flownet3d,gu2019hplflownet} (the cameras in the KITTI setup are mounted at $\approx1.65$~m height above ground). In \cref{tab:flow_traditional_supplement} we additionally report the results without removing the ground points. 

\parahead{lidarKITTI~\cite{menze2015joint,menze2018object}} This dataset comprises the Velodyne 64-beam LiDAR point clouds corresponding to the same 142 scenes used in \emph{stereoKITTI}. The ground truth scene flow vectors are obtained by projecting the points of the source point cloud to the image plane using the provided calibration parameters, and associating them with the scene flow vectors of the corresponding pixels from the \emph{stereoKITTI} dataset~\cite{liu2019meteornet}. Due to directly using the LiDAR point clouds the points of both frames are not in direct correspondence and exhibit the typical LiDAR sampling pattern with very uneven point density.
\begin{table}[!t]
    \setlength{\tabcolsep}{4pt}
    \renewcommand{\arraystretch}{1.25}
	\centering
	\resizebox{\columnwidth}{!}{
    \begin{tabular}{clccccc}
            \toprule
			Dataset & Method & Supervision & EPE3D [m]~$\downarrow$ & Acc3DS~$\uparrow$ & Acc3DR~$\uparrow$ & Outliers~$\downarrow$ \\
			\hline
			\multirow{6}{*}{\begin{tabular}{c}\emph{stereoKITTI}\\ (w/o ground) \end{tabular}}
            & HPLFlowNet~\cite{gu2019hplflownet} & Full & 0.117 & 0.478 & 0.778 & 0.410 \\
			& PointPWC-Net~\cite{wu2019pointpwc} & Full & 0.069 & 0.728 & 0.888 & 0.265 \\
			& FLOT~\cite{puy20flot} & Full & 0.056 & 0.755 & 0.908 & 0.242 \\
			& Ours (backbone)& Full & \textbf{0.042} & \textbf{0.849} & \textbf{0.959} & \textbf{0.208} \\
			\cline{2-7}
			& Ours & Weak & 0.163 & 0.541 & 0.658 & 0.452 \\
			& Ours++ & Weak & 0.134 & 0.709 & 0.800 & 0.311 \\
            \hline
			\multirow{5}{*}{\begin{tabular}{c}\emph{stereoKITTI}\\ (with ground) \end{tabular}}
            & HPLFlowNet~\cite{gu2019hplflownet} & Full & 0.238 & 0.194 & 0.429 & 0.787 \\
			& PointPWC-Net~\cite{wu2019pointpwc} & Full & 0.204 & 0.292 & 0.556 & 0.645 \\
			& FLOT~\cite{puy20flot} & Full & 0.122 & 0.480 & 0.691 & 0.401 \\
			& Ours (backbone) & Full & 0.143 & 0.392 & 0.660 & 0.533 \\
			\cline{2-7}
			& Ours & Weak & 0.136 & 0.470 & 0.712 & 0.420 \\
			& Ours++ & Weak & \textbf{0.068} & \textbf{0.836} & \textbf{0.897} & \textbf{0.263} \\
			\bottomrule
	\end{tabular}
	}
	\vspace{-1mm}
	\caption{Evaluation results on \emph{stereoKITTI} dataset. Weakly supervised models are trained on \emph{semanticKITTI} LiDAR point clouds and evaluated directly on stereo point clouds of \emph{stereoKITTI}. Ours and Ours++ denote the output of the network before and after test-time optimization, respectively.}
	\vspace{-4mm}
	\label{tab:flow_traditional_supplement}
\end{table}
\begin{table*}[!t]
    \setlength{\tabcolsep}{4pt}
    \renewcommand{\arraystretch}{1.15}
	\centering
	\resizebox{\textwidth}{!}{
    \begin{tabular}{ccc|cccc|cccc|cc}
            \toprule
            &&&\multicolumn{4}{c|}{scene flow}&\multicolumn{4}{c|}{BG - segmentation}&\multicolumn{2}{c}{ego motion}\\
			Dataset & Method & Supervision & mean EPE3D [m]~$\downarrow$ & med. EPE3D [m]~$\downarrow$ & med. F-EPE3D [m]~$\downarrow$ & med. B-EPE3D [m]~$\downarrow$  & prec. FG~$\uparrow$& rec. FG~$\uparrow$ & prec. BG~$\uparrow$ & recall. BG~$\uparrow$ & RRE [$^\circ$]~$\downarrow$ & RTE [m]~$\downarrow$ \\
            \hline
            \multirow{3}{*}{\begin{tabular}{c}\emph{lidarKITTI}\\ (w/o ground) \end{tabular}} 
			& Ours & Weak & 0.150 & 0.111 & 0.227 & 0.104  & 0.726 & 0.885 & 0.978 & 0.940  & 0.379  & 0.130\\
			& Ours+ & Weak & 0.110 & 0.064 & 0.227 & 0.049  & 0.726 & 0.885 & 0.978 & 0.940  & 0.126  & 0.053\\
			
			& Ours++ & Weak & 0.094 & 0.051 & 0.164 & 0.049  & 0.726 & 0.885 & 0.978 & 0.940  & 0.126  & 0.053\\
			\hline
			\multirow{3}{*}{\begin{tabular}{c}\emph{lidarKITTI}\\ (with ground) \end{tabular}} 
			& Ours & Weak & 0.133 & 0.109 & 0.186 & 0.109 & 0.734 & 0.855 & 0.991 & 0.980  & 0.327  & 0.130\\
			& Ours+ & Weak & 0.106 & 0.083 & 0.186 & 0.083 & 0.734 & 0.855 & 0.991 & 0.980  & 0.142 & 0.091\\
			
			& Ours++ & Weak & 0.103 & 0.083 & 0.131 & 0.083 & 0.734 & 0.855 & 0.991 & 0.980  & 0.142  & 0.091\\
			\bottomrule
	\end{tabular}
	}
	\vspace{0.3mm}
	\caption{Detailed results of out weakly supervised method on \emph{lidarKITTI} dataset. Ours denotes the direct output of the network. Ours+ and Ours++ are the results with only background and full test-time optimization, respectively}
	\label{tab:kitti_lidar_supplementary}
	\vspace{-4mm}
\end{table*}

\parahead{semanticKITTI~\cite{behley2019iccv}} This dataset provides pointwise semantic labels annotated in 3D and improved ego-motion information for all sequences of the KITTI odometry benchmark~\cite{Geiger2012CVPR}. Different to \emph{stereoKITTI} and \emph{lidarKITTI}, which include only the points that map to the image plane of the front camera, \emph{semanticKITTI} contains full $360^\circ$ LiDAR sweeps. We therefore process \emph{semanticKITTI} point clouds to make them consistent with \emph{lidarKITTI}. Specifically, we first convert the 3D coordinates of the points to polar coordinates and consider only the points with an azimuth angle in the range $[-45^\circ, 45^\circ]$ and elevation angle in the range $[-24.9^\circ, 12.0^\circ]$. This maintains the points that would roughly map to the image plane of the front camera. We then additionally remove the points that are less than $1.5$~m or more than $35$~m away from the LiDAR sensor. The binary background mask is generated by combining the semantic labels into the background (class labels from 40 to 249 ) and foreground (other class labels)\footnote{\url{https://github.com/PRBonn/semantic-kitti-api/blob/master/config/semantic-kitti-all.yaml}}. We split the point cloud pairs of \emph{semanticKITTI} into 4350 validation (sequences 03 and 05) and 18840 training samples (remaining nine sequences of the training set). Because BG-FG labels of the test dataset are not available, we perform all evaluations on the validation dataset.

\parahead{waymo open~\cite{sun2019scalability}} This is a large scale dataset collected by a fleet of waymo self-driving cars in various conditions. It contains 1950 sequences of $20$~s duration each, collected with an acquisition rate of 10\,Hz. In our experiments we use the training batches 0--16 ($\approx50\%$ of the training data) for fine-tuning (\cref{sec:additional_results}) and the validation batches 0--2 ($\approx40\%$ of the validation data) for evaluating our approach\footnote{\url{https://waymo.com/open/download/}}. Waymo cars are equipped with five LiDAR sensors altogether, one mid-range LiDAR on the top and four short range ones on the sides of the car. In our experiments, we only use the points acquired by the top, mid-range scanner. We then transform these points into a cordinate system centered at the location of the LiDAR sensor in the KITTI setup and follow the processing steps used in \emph{semanticKITTI} dataset to extract only the points that would roughly project to the front camera. Along with the ego-motion information, \emph{waymo-open} also provides the 3D bounding-boxes of vehicles, pedestrians, cyclists, and signs. We use the former three classes to extract the foreground and consider the remaining points as background.
\begin{figure*}[ht]
    \centering
    \includegraphics[width=0.96\linewidth]{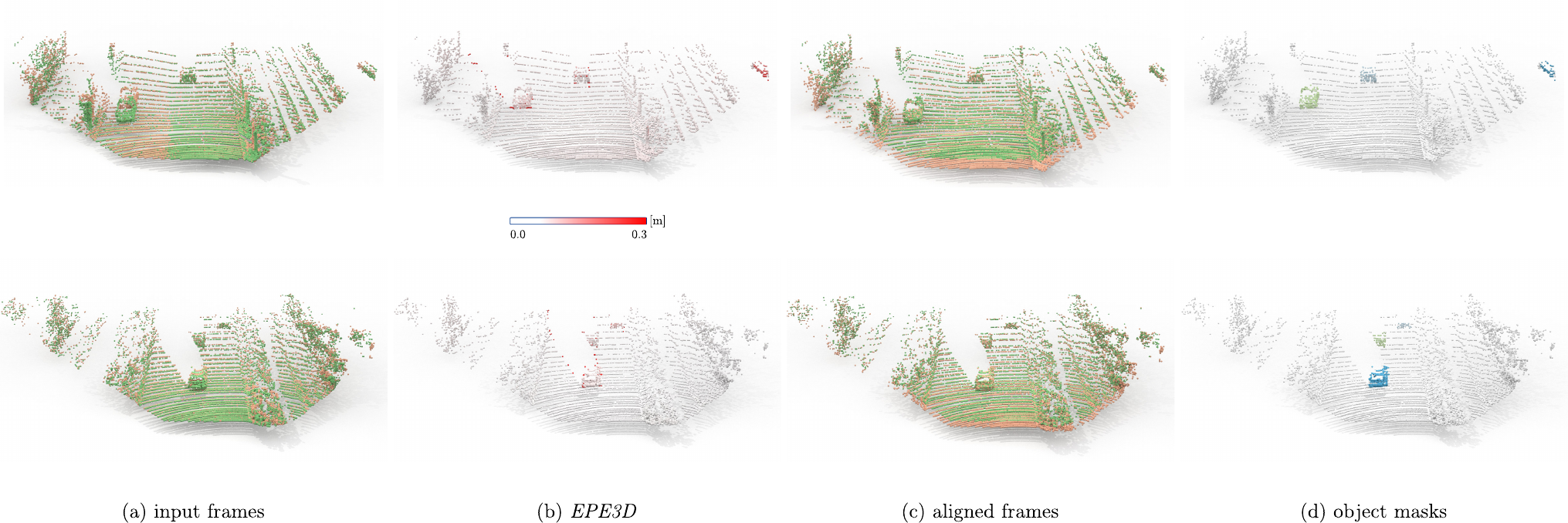}
    \caption{Successful cases of our method on the \emph{lidarKITTI} dataset. By correctly splitting the scene into foreground and background (d), our method estimates the accurate scene flow vectors (b), which align the two frames (c).}
    \label{fig:good_cases_lidar_kitti}
    \vspace*{0.5\floatsep}
    \includegraphics[width=0.96\linewidth]{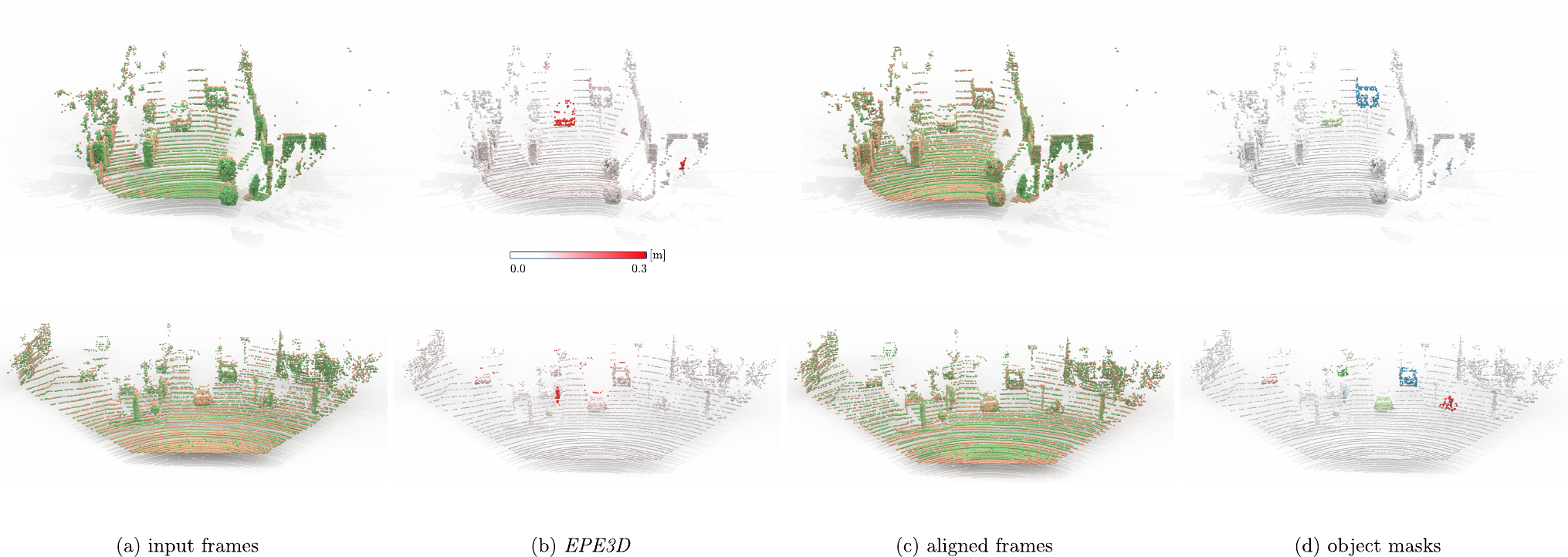}
    \caption{Failure cases of our method on the \emph{lidarKITTI} dataset. \textbf{Top:} even though the car's object mask (d) is correctly predicted, its predicted scene flow vectors yield large end-point-errors (b). \textbf{Bottom:} a pillar in the middle of the scene is wrongly predicted as foreground object (d), hence its scene flow does not agree with the background and GT (b). %
    }
    \label{fig:failure_cases_lidar_kitti}
    \vspace*{0.5\floatsep}
    \includegraphics[width=0.96\linewidth]{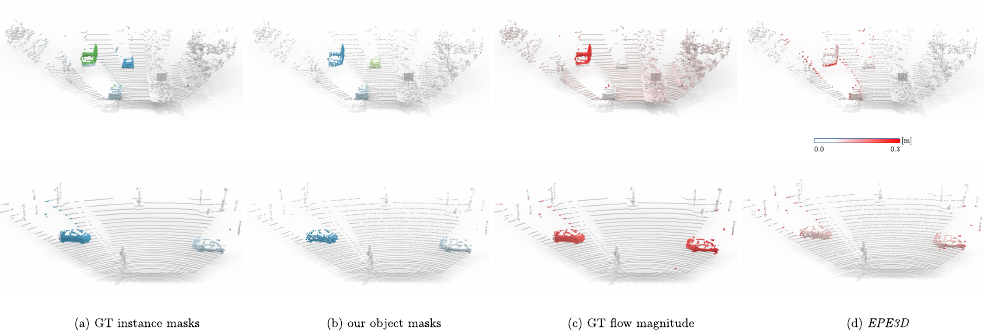}
    \caption{\emph{lidarKITTI} annotations are obtained by projecting 3D points onto the image plane, which results in wrong instance (a) and scene flow (c) annotations for points with azimuth and elevation angles close to the object boundaries (e.g. green car is partially blue in (a) top). On the other hand, our method infers correct object masks (b) and scene flow (d), yet due to the wrong GT annotations, the scene flow appears to be erroneous.}
    \label{fig:annotation_problems}
\end{figure*}

\subsection{Evaluation details and additional results}
\label{sec:evaluations}
We start this section by providing additional details and results (\cref{sec:experimental_details}) supporting the evaluations presented in \cref{sec:experiments}, before reporting additional evaluations (\cref{sec:additional_results}) that were omitted from the main paper due to the space constraint.

\vspace{-1mm}
\subsubsection{Evaluation details}
\label{sec:experimental_details}
\vspace{-1mm}
\parahead{Additional results on \emph{stereoKITTI}}
In \cref{sec:experiments_backbone}, we evaluate the performance of our backbone under full supervision on \emph{FlyingThings3D} and \emph{stereoKITTI}. \cref{tab:flow_traditional_supplement} supplements that section with the evaluation results on the \emph{stereoKITTI} dataset without removing the ground points. Additionally, we report the generalization results of our weakly supervised model trained on the LiDAR point clouds of \emph{semanticKITTI}.
 
As expected, the performance of all fully supervised methods drops significantly if the challenging ground points are not removed. Remarkably, our weakly supervised model generalizes from LiDAR to stereo point clouds, and when combined with the test-time optimization even outperforms all methods on \emph{stereoKITTI} with ground points. \cref{tab:flow_traditional_supplement} also hints that the generalization LiDAR $\mapsto$ stereo is less challenging than the opposite stereo $\mapsto$ LiDAR.

\begin{table}[!t]
    \setlength{\tabcolsep}{4pt}
    \renewcommand{\arraystretch}{1.15}
	\centering
	\resizebox{\columnwidth}{!}{
    \begin{tabular}{ccccccc}
            \toprule
			Dataset & Method & Initialization & EPE3D [m]~$\downarrow$ & Acc3DS~$\uparrow$ & Acc3DR~$\uparrow$ & Outliers~$\downarrow$ \\
            \hline
            \multirow{4}{*}{\begin{tabular}{c}\emph{lidarKITTI}\\ (w/o ground) \end{tabular}} 
            
			& Ours & Pretrained & 0.102 & 0.706 & 0.833 & 0.357 \\
			& Ours++ & Pretrained & \textbf{0.080} & \textbf{0.834} & \textbf{0.912} & \textbf{0.279} \\
			& Ours & Random & 0.150 & 0.521 & 0.744  & 0.450 \\ 
			& Ours++ & Random & 0.094 & 0.784 & 0.885 & 0.314 \\
			\hline
			\multirow{4}{*}{\begin{tabular}{c}\emph{lidarKITTI}\\ (with ground) \end{tabular}} 
			& Ours   & Pretrained & 0.091 & 0.601  & 0.788 & 0.445 \\
			& Ours++ & Pretrained &  \textbf{0.080} &  \textbf{0.742}  & \textbf{0.850} &  \textbf{0.369} \\
			& Ours   & Random &  0.133 &  0.460 & 0.746 & 0.527 \\
		    & Ours++ & Random & 0.103  & 0.686  & 0.819 & 0.410 \\
			\bottomrule
	\end{tabular}

	}
	\caption{Evaluation of our model trained with random initialization of the weights (Random) and with the weights pretrained on \emph{FT3D} (Pretrained), on \emph{lidarKITTI} dataset. Ours and Ours++ denote the direct output of the network and the result after test-time optimization, respectively.}
	\label{tab:ablation_pretraining}
	\vspace{-3mm}
\end{table}

\parahead{Additional results on \emph{lidarKITTI}}
\cref{tab:kitti_lidar_supplementary} supplements \cref{tab:kitti_lidar_flow} and provides detailed results of our method on the \emph{lidarKITTI} dataset. Note, how the test-time optimization improves background (Ours+) as well as foreground (Ours++) scene flow estimates. The results of the BG-segmentation and FG/BG scene flow split should be interpreted with caution, due to the noisy annotation of the \emph{lidarKITTI} dataset (see below).
Further qualitative results are shown in \cref{fig:good_cases_lidar_kitti} and failure cases in \cref{fig:failure_cases_lidar_kitti}

\begin{table}[!t]
    \setlength{\tabcolsep}{4pt}
    \renewcommand{\arraystretch}{1.2}
	\centering
	\resizebox{\columnwidth}{!}{
    \begin{tabular}{cc|cccc|cc}
            \hline
             & & \multicolumn{4}{c|}{BG segmentation} & \multicolumn{2}{c}{ego-motion}\\
            
             Method & Initialization  & prec. FG~$\uparrow$& rec. FG~$\uparrow$ & prec. BG~$\uparrow$ & recall. BG~$\uparrow$ & RRE [$^\circ$]~$\downarrow$ & RTE [m]~$\downarrow$ \\
            \hline
            \multicolumn{8}{c}{\textit{semanticKITTI} (w/o ground)}\\
            \hline
            Ours  & Pretrained  & 0.950 & 0.892 & \textbf{0.991} & 0.996 & 0.145 & 0.035 \\
            Ours++  & Pretrained  & 0.950 & 0.892 & \textbf{0.991} & 0.996 & \textbf{0.127} & \textbf{0.031} \\
            Ours  & Random & \textbf{0.971} & \textbf{0.895} & \textbf{0.991} & \textbf{0.998} & 0.201 & 0.047 \\
            Ours++  & Random &  \textbf{0.971} & \textbf{0.895} & \textbf{0.991} & \textbf{0.998} & 0.133 &  0.032  \\            
            \hline
             \multicolumn{8}{c}{\textit{semanticKITTI} (with ground)}\\
            \hline
            Ours  & Pretrained & 0.942 & \textbf{0.909} & \textbf{0.996} & 0.998 & 0.177 & 0.044 \\
            Ours++  & Pretrained  & 0.942 & \textbf{0.909} & \textbf{0.996} & 0.998 & \textbf{0.116} & \textbf{0.029} \\
            Ours  & Random & \textbf{0.966} & 0.904 & \textbf{0.996} & \textbf{0.999} & 0.249 & 0.059 \\
            Ours++  & Random  & \textbf{0.966} & 0.904 & \textbf{0.996} & \textbf{0.999} & 0.121 & 0.032 \\      
			\hline
	\end{tabular}
	}
	\vspace{0.1mm}
	\caption{Evaluation of our model trained with random initialization of the weights (Random) and with the weights pretrained on \emph{FT3D} (Pretrained), on \emph{semanticKITTI} dataset. Ours and Ours++ denote the direct output of the network and the result after test-time optimization, respectively.}
	\label{tab:ablation_pretraining_semantickitti}
	\vspace{-6mm}
\end{table}
\parahead{Noisy annotations of \emph{lidarKITTI}}
\label{sec:annottation_problems}
During the evaluation, we have discovered that \emph{lidarKITTI} annotations are noisy and contain outliers. These GT errors occur especially around the points lying on or close to the object boundaries.
Outliers are caused by the projection of the 3D point onto the 2D image plane, where two distant points in 3D can map to the same pixel, as there is no perception of depth. \cref{fig:annotation_problems} shows two prominent examples: in some cases, the instance mask and motion of one object are also assigned to the other (top), and in other cases, the background points get assigned the instance label and scene flow of the object in the front (bottom). Because of training on \emph{semanticKITTI} with accurate annotations, our method is still capable of predicting the correct object masks (\cref{fig:annotation_problems} (b)). Wrong annotation however result in a lower BG-segmentation performance and cause apparent errors in our scene flow prediction (see \cref{fig:annotation_problems} (c)). 
Unfortunately, this error is reflected in the quantitative evaluations not only for our method but also for many other scene flow algorithms out there.
To reduce this effect, we additionally report median \emph{EPE3D} values in \cref{tab:kitti_lidar_supplementary}, as well as the FG and BG \emph{EPE3D} based on our predicted BG-mask, instead of the noisy GT mask.
\vspace{-1mm}
\subsubsection{Additional evaluations and ablation studies}
\label{sec:additional_results}
\vspace{-2mm}

\parahead{Pretraining vs training from scratch}
Evaluations reported in \cref{sec:experiments_weakly_supervised} to \cref{sec:ablation} are performed using the weakly supervised model trained on \emph{semanticKITTI} from scratch. However, recent works show that general 3D backbone networks can benefit from pretraining on large (annotated) datasets~\cite{xie2020pointcontrast}. 

To evaluate this in our setting, we consider a model whose backbone weights were initialized with weights trained (with full supervision) on \emph{FlyingThings3D} rather than randomly. The evaluations on \emph{lidarKITTI} (\cref{tab:ablation_pretraining}) and \emph{semanticKITTI} (\cref{tab:ablation_pretraining_semantickitti}) show that, in line with the literature, our backbone network can indeed benefit from pretraining. The improvement in is especially prominent in scene flow estimation (more than $1$~cm lower \emph{EPE3D} in \cref{tab:ablation_pretraining}) and in ego-motion estimation (lower rotation and translation errors in \cref{tab:ablation_pretraining_semantickitti}).
In order to fully adhere to the weakly supervised setting, we use the randomly initialized model in all other evaluations.

\begin{figure*}[ht]
  \centering
    \includegraphics[width=0.99\linewidth]{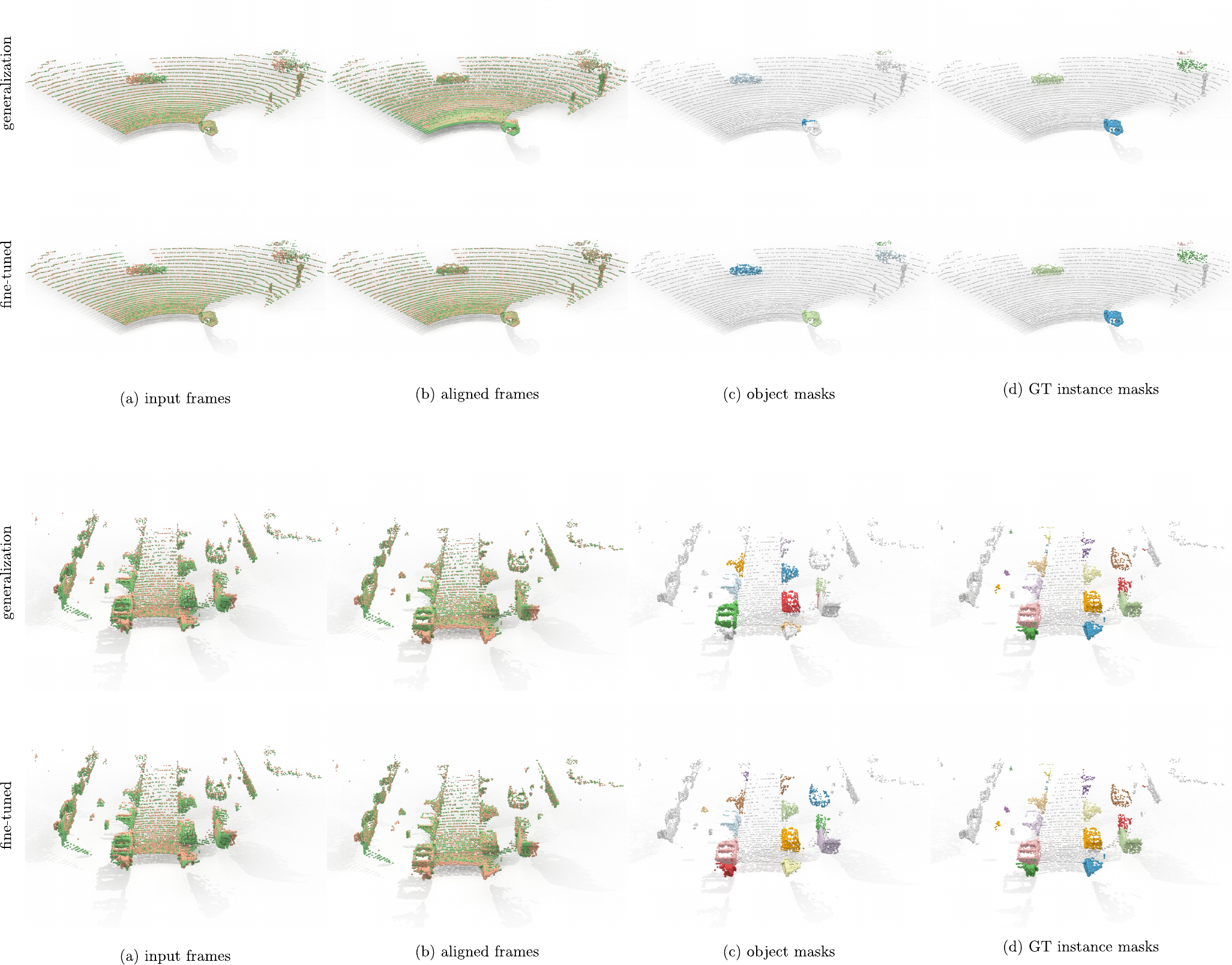}
    \caption{Fine-tuning on \emph{waymo open} improves performance and robustifies our model. \textbf{Top:} objects close to the sensor are not common in \emph{semanticKITTI} and hence cannot be detected correctly by the generalized model (c). \textbf{Bottom:} a challenging example with 18 foreground objects (much larger than average number in \emph{semanticKITTI}). Note, how more object masks are correctly inferred by our fine-tuned model compared to direct generalization (c-column). %
    }
    \label{fig:fine_tuning_waymo}
    \vspace*{0.6\floatsep}
    \includegraphics[width=0.99\linewidth]{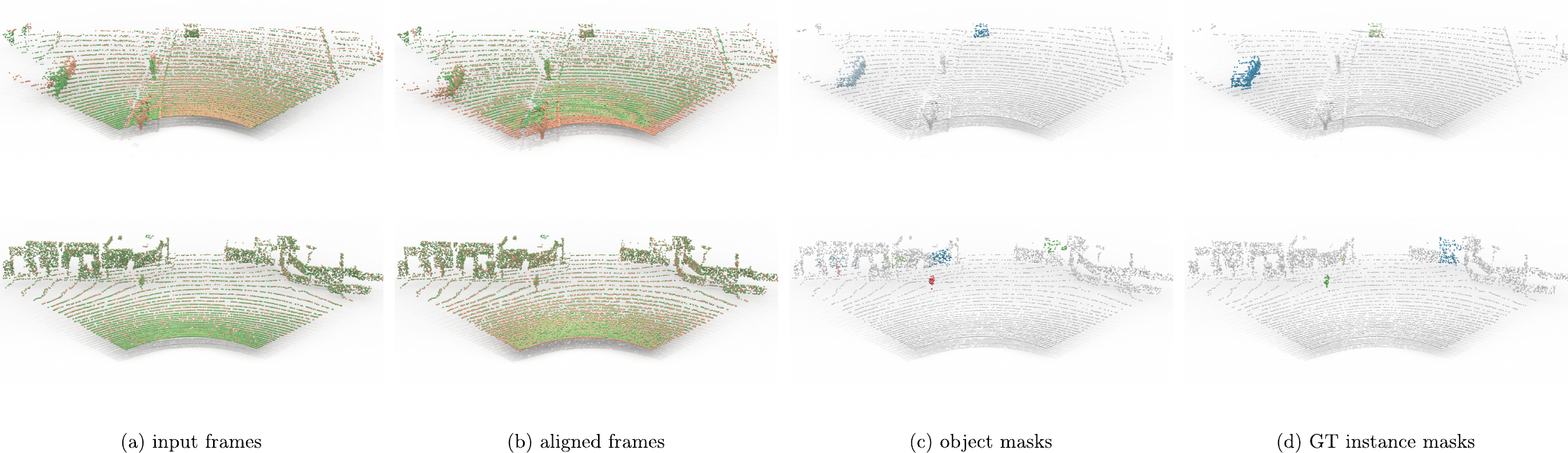}
    \caption{Failure cases on \emph{waymo open} dataset. \textbf{Top:} our model is unable to estimate accurate ego-motion and scene flow (b) if the background points consists only of the ground points after foreground removal (c). \textbf{Bottom:} rare objects such as trucks (top right corner in c and d) appear ambiguous to our model and cause prediction of the wrong masks (c). %
    }
    \label{fig:failed_cases_waymo}
\end{figure*}

\parahead{Fine-tuning on \emph{waymo open}}
\begin{table}[!t]
    \setlength{\tabcolsep}{4pt}
    \renewcommand{\arraystretch}{1.2}
	\centering
	\resizebox{\columnwidth}{!}{
    \begin{tabular}{cc|cccc|cc}
            \hline
             & & \multicolumn{4}{c|}{BG segmentation} & \multicolumn{2}{c}{ego-motion}\\
            
             Method & Mode  & prec. FG~$\uparrow$& rec. FG~$\uparrow$ & prec. BG~$\uparrow$ & recall. BG~$\uparrow$ & RRE [$^\circ$]~$\downarrow$ & RTE [m]~$\downarrow$ \\
            \hline
            \hline
            Ours++  & generalization  &  \textbf{0.960} & 0.689 & 0.957 & \textbf{0.996} & 0.141 & 0.099 \\
            Ours++  & fine-tuned & 0.945 & \textbf{0.921} & \textbf{0.989} & 0.992 & \textbf{0.111} & \textbf{0.078} \\
			\hline
	\end{tabular}
	}
	\caption{Comparison of the model fine-tuned on \emph{waymo open} with the model trained only on \emph{semantiKITTI} (generalization), on the \emph{waymo open} dataset. Fine-tuned model outperforms the directly generalized one in terms of FG precision and ego-motion error.}
	\vspace{-5mm}
	\label{tab:ablation_finetuning_waymoopen}
\end{table}
In the main paper, \emph{waymo open} dataset is used only to evaluate the direct generalization performance of our model trained on \emph{semanticKITTI}. However, as briefly discussed in the main paper, \emph{waymo  open} also provides all the annotations that our weakly supervised model relies on. We now use these annotations, to evaluate the gain obtained by fine-tuning our model. To this end, we initialize our model with the weights trained on \emph{semanticKITTI} and fine-tune it for 22k iterations (less than 2 epochs) on \emph{waymo open}. \cref{tab:ablation_finetuning_waymoopen} shows that the fine-tuned model greatly outperforms the model trained only on \emph{semanticKITTI}, especially in terms of foreground recall (gain of more than 20 percent points) and relative translation error (improvement of $2$~cm)%
. The relative improvement is  qualitatively also depicted in \cref{fig:fine_tuning_waymo}

\parahead{Training and testing with GT instance masks}
\begin{table}[!t]
    \setlength{\tabcolsep}{6pt}
    \renewcommand{\arraystretch}{1.15}
	\centering
	\resizebox{\columnwidth}{!}{
    \begin{tabular}{ccccccc}
			\hline
			\multicolumn{2}{c}{GT inst. mask} & \multicolumn{5}{c}{\textit{lidarKITTI} (with ground)} \\
            \cline{3-7}
            \multicolumn{1}{c}{train} & \multicolumn{1}{c}{test}  & EPE [m]~$\downarrow$ & F-EPE [m]~$\downarrow$ & B-EPE [m]~$\downarrow$ & RRE [$^\circ$]~$\downarrow$ & RTE [m]~$\downarrow$ \\
            \hline
            \multicolumn{1}{c}{\ding{51}} & \multicolumn{1}{c}{}  & \textbf{0.097} & 0.216 & \textbf{0.085} & 0.146 & \textbf{0.082}\\
            
            \multicolumn{1}{c}{} & \multicolumn{1}{c}{\ding{51}}  &  0.101 & \textbf{0.183} & 0.094 & \textbf{0.139} & 0.091\\
            
            \multicolumn{1}{c}{\ding{51}} & \multicolumn{1}{c}{\ding{51}} &  \textbf{0.097} &  0.265 & \textbf{0.085} & 0.146 & \textbf{0.082} \\
            
             &  &   0.102 & 0.195 & 0.094 & \textbf{0.139} & 0.091\\
			\hline
	\end{tabular}
	}
	\caption{Performance evaluation of our simple foreground clustering algorithm compared to models using GT instance mask during training and/or testing on \emph{lidarKITTI} dataset.}
	\label{tab:ablation_gt_clusters}
\end{table}
To define the FG instance-level rigidity loss and to perform the test-time optimization of the foreground, we rely on a simple unsupervised clustering of the foreground into individual objects. Arguably, this part could be replaced by a more powerful instance segmentation head, which would however require instance annotations during training. It is of interest to see how much benefit a \emph{perfect} clustering would bring to our method. We therefore ablate by replacing the output of our clustering with the GT instance labels, which are provided both in \emph{semanticKITTI} and \emph{lidarKITTI}. %
We assess the individual contributions from having GT labels during training, testing or both.
\cref{tab:ablation_gt_clusters} shows that our model with the simple clustering algorithm performs comparably to the models using GT instance masks during training (\emph{semanticKITTI})  and/or testing (\emph{lidarKITTI}). We conclude that our
clustering algorithm, which does not require GT instance masks during training or testing, is the preferred option. Due to the noisy instance annotations of the \emph{lidarKITTI} dataset, the performance of models using GT instance masks during testing should be interpreted with caution.

\parahead{Sinkhorn algorithm}
\begin{table}[!t]
    \setlength{\tabcolsep}{18pt}
    \renewcommand{\arraystretch}{1.05}
	\centering
	\resizebox{\columnwidth}{!}{
    \begin{tabular}{cccc}
			\hline
			\multicolumn{1}{c}{} & \multicolumn{3}{c}{\textit{lidarKITTI} (with ground)} \\
            \cline{2-4}
            & EPE [m]~$\downarrow$ & RRE [$^\circ$]~$\downarrow$ & RTE [m]~$\downarrow$ \\
            \hline
            \multicolumn{1}{c}{w/o Sinkhorn} & 0.594 & 0.539 & 0.615\\
            \multicolumn{1}{c}{with Sinkhorn} & \textbf{0.133} & \textbf{0.327} & \textbf{0.130}  \\
			\hline
	\end{tabular}
	}
	\vspace{0mm}
	\caption{Ablation study of the Sinkhorn algorithm on \emph{lidarKITTI} dataset (evaluation protocol follows Tab. 3 in the main paper).\vspace{-6mm}} 
	\label{tab:ablation_sinkhorn}
\end{table}
\review{The benefit of using the entropy-regularized Sinkhorn algorithm for the ego-motion estimation is two fold: (i) the Sinkhorn distances should yield more accurate correspondences due to the optimally of the transport map, and (ii) in combination with the slack row and column it enables us to down-weight the outliers in a principled manner. We empirically confirm these benefits in an ablation study in which we directly use the affinity matrix $\mathbf{M}$ to compute the soft correspondences $\phi(\mathbf{x}_i^b,\mathbf{Y}^b) = \mathbf{Y}^b\mathbf{m}_i/||\mathbf{m}_i||_1$ and estimate the ego-motion with an unweighted Kabsch algorithm (\cref{sec:kabsch}).  

\cref{tab:ablation_sinkhorn} depicts a significant increase of the EPE when Sinkhorn is deactivated. This increase can be accredited to the inferior estimation of the ego-motion parameters, which reults in a much higher RTE and RRE.}

\end{document}